%% file: main.tex
\documentclass[conference]{ieeeconf}
\usepackage{amsmath}
\usepackage{amssymb}
\usepackage{mathtools}
\usepackage{wasysym}
\usepackage{epstopdf}
\usepackage{graphicx}
\usepackage{subcaption}
\usepackage{booktabs}
\usepackage{xcolor}
\usepackage{placeins}
\usepackage{makecell}
\usepackage{arydshln}

\usepackage{algorithm}          % float wrapper
\usepackage{algpseudocode}      % provides \Require, \Ensure, \State, \While, etc.
\usepackage{todonotes}

\usepackage[hidelinks]{hyperref}

\title{Scalable Surface-Based Manipulation Through Modularity and Inter-Module Object Transfer}
\author{
    Pratik Ingle, Jørn Lambertsen, Kasper Støy, Andres Faiña\\
    \small IT University, Denmark\\ 
    \small \{prin, jrnl, ksty, anfv\}@itu.dk
}

\begin{document}
\maketitle

% Heterogeneous object manipulation on nonlinear soft surface through linear controller}}
% \author{
%     Pratik Ingle, Jørn Lambertsen, Kasper Støy, Andres Faiña\\
%     \small IT University, Denmark\\ 
%     \small \{prin, jrnl, ksty, anfv\}@itu.dk
% }
\maketitle

\begin{abstract}

Robotic Manipulation Surfaces (RMS) manipulate objects by
deforming the surface on which they rest, offering safe,
parallel handling of diverse and fragile items. However,
existing designs face a fundamental tradeoff: achieving fine
control typically demands dense actuator arrays that limit
scalability. Modular architectures can extend the workspace, but
transferring objects reliably across module boundaries on soft,
continuously deformable surfaces remains an open challenge. We
present a multi-modular soft manipulation platform that achieves
coordinated inter-module object transfer and precise positioning
across interconnected fabric-based modules. A
hierarchical control framework, combining conflict-free
Manhattan-based path planning with directional object passing
and a geometric PID controller, achieves sub-centimeter
positioning and consistent transfer of heterogeneous objects
including fragile items. The platform employs shared-boundary
actuation, where adjacent modules share edge actuators, reducing
the required count from $4n^2$ to $(n+1)^2$ for an
$n \times n$ grid; a 2$\times$2 prototype covers
1$\times$1~m with only 9 actuators. This scaling comes at a
cost: shared actuators mechanically couple neighbouring modules,
creating interference during simultaneous manipulation. We
systematically characterise this coupling across spatial
configurations and propose compensation strategies that reduce
passive-object displacement by 59--78\%. Together, these
contributions establish a scalable foundation for soft
manipulation surfaces in applications such as food processing
and logistics.

\end{abstract}

\input{introduction}

\input{sms}

\input{method}
\input{results}
\input{Discussion}

\input{conclusion}
\bibliographystyle{unsrt}
\bibliography{ref}
\vspace{12pt}
\end{document}

%% file: introduction.tex
\section{Introduction}

\begin{figure}[!t]
    \centering
    % First row with three columns
    \begin{subfigure}{0.5\textwidth}
        \centering
        \includegraphics[width=\textwidth]{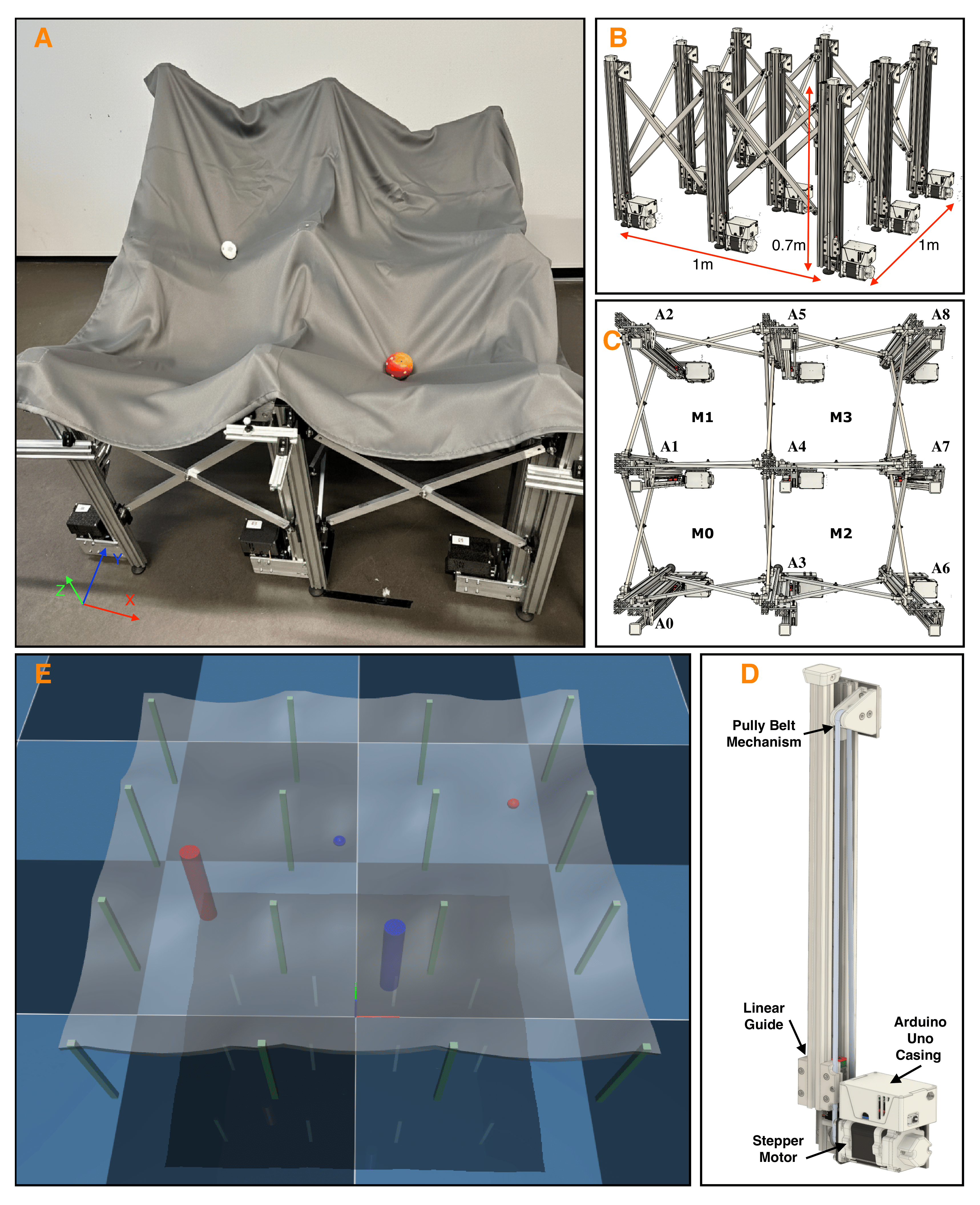} 
        % \caption{Side view of the platform without the flexible surface}
        \label{fig:side mantaray}
    \end{subfigure}
    % \caption{\textbf{Multi-Modular MANTA-RAY:} Manipulation with Adaptive Non-rigid Textile Actuation with Reduced Actuation densitY platform. (A) Manipulation of an egg and an apple. (B) Side view showing the platform dimensions of $1 \times 1 \times 0.7$~m (length, width, and height at rest). (C) Top view illustrating the arrangement of nine actuators (A0–A8) and four modules (M0–M3) without the flexible surface. (D) Each linear actuator has 0.4-meter vertical range and built using a stepper motor and pulley-belt mechanism, controlled by an \textit{Arduino Uno} micro controller. (E) MuJoCo simulation of the extended $3\times3$-module configuration showing two objects (red and blue spheres) and their respective target locations.}
    \caption{Multi-modular soft manipulation. (A) Manipulation of an egg and an apple. (B) Side view showing the platform dimensions of $1 \times 1 \times 0.7$~m (length, width, and height at rest). (C) Top view illustrating the arrangement of nine actuators (A0–A8) and four modules (M0–M3) without the flexible surface. (D) Each linear actuator has 0.4-meter vertical range and built using a stepper motor and pulley-belt mechanism, controlled by an \textit{Arduino Uno} micro controller. (E) MuJoCo simulation of the extended $3\times3$-module configuration showing two objects (red and blue spheres) and their respective target locations.}
    % \caption{multi-modular MANTA-RAY : multi modular Manipulation with Adaptive Non-rigid Textile Actuation with Reduced Actuation densitY platform (A) manipulating an egg and apple. Side view (B) shows the dimensions of the platform, $1 \times 1 \times 0.7$ meter length, width and height at rest position. Top view (C) shows the orientation of 9 actuators (A0-A8) and four modules (M0-M3) of the platform without flexible surface. Each linear actuator (D) is build with stepper motor and pully belt mechanism, control with ardunio uno micro controller. MuJoCo simulation use to extend the platform to further $3\times3 module$ (E), Two objects (red and blue sphere) and their respective target location}
    \label{fig:mutli-modular MANTA-RAY}
\end{figure}

In domains such as food processing, logistics, and warehouse automation, large numbers of heterogeneous and often fragile objects must be repositioned across extended work areas. Pick-and-place manipulators address this by grasping objects sequentially, but they risk damaging delicate items, struggle with irregular geometries, and are inherently limited in throughput when many objects must move simultaneously. Conveyor belts offer high throughput but are limited to unidirectional transport and cannot reposition objects to arbitrary locations. Robotic Manipulation Surfaces (RMS) offer a fundamentally different strategy: rather than grasping objects directly, they manipulate them by dynamically deforming or actuating the surface on which the objects rest. The distributed contact provided by such surfaces enables safe, parallel handling of diverse objects; including fragile items like eggs and fruits without requiring object-specific end-effectors. These properties make RMS especially attractive for applications that demand high throughput, gentle handling, and operation over large surface areas. However, realizing these advantages at practical scales remains an open challenge, as existing RMS designs face a fundamental tension between manipulation capability and actuator complexity.

A variety of RMS architectures have been proposed, each
navigating this tradeoff differently. Wheel-based systems \cite{uriarte2022methode} use arrays of omnidirectional rollers to transport objects efficiently but require continuous contact and struggle with small or geometrically irregular items that cannot maintain stable engagement with the rollers. Piston-based arrays \cite{johnson2023multifunctional} offer precise, localized control by independently actuating dense grids of vertical pins, enabling both translation and rotation of objects. However, their fine-grained manipulation comes at the cost of high actuator density. ArrayBot \cite{xue2024arraybot}, for instance, employs a 16×16 array of 256 actuators which increases mechanical complexity, fabrication cost, and control overhead as the workspace grows. Alternative approaches based on vibrating plates, airflow levitation, cilia-inspired actuators and deformable gel layers \cite{zhou2016controlling, moon2006distributed} each exploit different physical principles for object transport, but remain constrained by trade-offs in payload capacity, object versatility, or the ability to scale to larger work areas. Across these designs, a common pattern emerges: achieving fine manipulation control typically demands high actuator density, which in turn limits practical scalability. Conversely, designs that reduce actuator count sacrifice the ability to handle diverse objects or achieve precise positioning. Breaking this coupling between manipulation capability and actuator complexity is essential for deploying RMS in real-world, large-area applications.

Chih-Han et al.\cite{yu2008morpho} took a step toward addressing this tradeoff with Morpho, a modular self-deformable robot that uses membrane-based surface deformation for inter-module object transfer, though its focus remained on global shape morphing rather than precise manipulation. More recent work, \cite{11020841} proposed MANTA-RAY a fabric-based manipulation platform that directly targets the density–capability coupling by using a soft polyester fabric surface supported by only four corner-mounted linear actuators. Despite this drastically reduced actuator count, a single module achieves an object-to-actuator size ratio as low as 0.01 and successfully manipulates fragile and heterogeneous objects such as eggs and fruits using a simple geometric transformation-based PID controller without requiring learning-based methods. However, these results have been demonstrated only on a single
module with a fixed workspace (0.5$\times$0.5~m). Extending to
multiple modules introduces two challenges that do not arise in
the single-module case: transferring objects reliably across
module boundaries where the continuous fabric surface deforms
unpredictably, and managing the mechanical coupling that arises
when neighbouring modules share physical actuators. Neither
challenge has been addressed for soft manipulation surfaces.

% Chih-Han et al.\cite{yu2008morpho} took a step toward addressing this tradeoff with Morpho, a modular self-deformable robot that uses membrane-based surface deformation for inter-module object transfer, though its focus remained on global shape morphing rather than precise manipulation. More recently, Ingle et al. \cite{11020841, 11163787} proposed the MANTA-RAY platform, which directly targets the density–capability coupling by using a soft polyester fabric surface supported by only four corner-mounted linear actuators. Despite this drastically reduced actuator count, a single module achieves an object-to-actuator size ratio as low as 0.01 and successfully manipulates fragile and heterogeneous objects such as eggs and fruits using a simple geometric transformation-based PID controller without requiring learning-based methods. However, these results have been demonstrated only on a single module with a fixed workspace ($0.5 \times 0.5 m$), and the feasibility of extending this approach to multi-module configurations where multiple objects must be manipulated simultaneously across interconnected modules remains unexplored.

Extending MANTA-RAY approach to multiple modules is not a straightforward replication of the single-module design. Naïvely placing independent four-actuator modules side by side would require 4n² actuators for an n×n grid, reintroducing the high density the original platform was designed to avoid. We instead propose a shared-actuator architecture where adjacent modules share edge actuators, reducing the count to (n+1)², yielding 9 instead of 16 actuators for a 2×2 grid (44\% reduction), with savings approaching 75\% asymptotically. However, this efficiency introduces mechanical coupling: shared actuators serve multiple modules simultaneously, constraining inter-module independence during parallel manipulation. Furthermore, transferring objects across module boundaries requires coordination strategies beyond the single-module controller. Addressing these coupled challenges demands both a compatible modular hardware design and a hierarchical control framework neither of which existed for soft manipulation surfaces prior to this work. We used a $2\times2$ module layout to study these effects in more detail. 

In this work, we present a multi-modular platform and make
three contributions: (1)~a hierarchical control framework that
achieves, reliable inter-module object
transfer on a soft deformable surface, combining conflict-free
Manhattan-based path planning with directional passing and
geometric PID control to enable seamless manipulation across
interconnected modules including parallel multi-object handling;
(2)~a systematic experimental characterisation of the
interference created by shared actuators, quantifying how
coupling scales with neighbourhood topology and proposing
compensation strategies that reduce passive-object displacement
by 59--78\%; and (3)~a shared-boundary hardware architecture in
which adjacent modules share edge actuators, achieving
$(n+1)^2$ actuator scaling and realised as a reconfigurable
2$\times$2 prototype covering 1$\times$1~m with only 9
actuators. We validate the system through hardware experiments
on seven heterogeneous objects and MuJoCo simulations on
3$\times$3 module configurations.

%% file: sms.tex
\section{Multi-Modular PLATFORM} \label{SMS_RAD}

% \begin{figure*}[ht]
%     \centering
%     % First row with three columns
%     \begin{subfigure}{1\textwidth}
%         \centering
%         \includegraphics[width=\textwidth]{figs/IMG_page-0001.jpg} 
%         % \caption{Side view of the platform without the flexible surface}
%         \label{fig:side mantaray}
%     \end{subfigure}
%     \caption{Muti-module MANTA-RAY platform (A) manipulating an egg and apple. Side view (B) and top view (C)  of the platform without flexible surface}
%     \label{fig:mutli-modular MANTA-RAY}
    
% \end{figure*}

\subsection{Design Concept and Scaling Architecture}

The single-module MANTA-RAY platform manipulates objects on a
soft fabric surface using four corner-mounted vertical linear
actuators that tilt and deform the surface to induce object
motion through three mechanisms: rolling, sliding, and pulling
\cite{11020841}. To extend this to larger workspaces, we
arrange actuators in an $(n+1) \times (n+1)$ grid, forming $n
\times n$ modules that share boundary actuators with their
neighbours. For a 2$\times$2 configuration (Fig.~\ref{fig:mutli-modular MANTA-RAY}C),
nine actuators (A0--A8) define four modules (M0--M3), where
adjacent modules share two actuators (e.g., A3 and A4 between
M0 and M2) and diagonal modules share one (A4 between M0 and
M3). This shared-boundary architecture requires only $(n+1)^2$
actuators instead of the $4n^2$ needed by independent modules,
reducing actuator count by 44\% at the 2$\times$2 scale with
savings approaching 75\% asymptotically. Each actuator operates
as an independent unit, providing robustness to individual
faults.

The multi-modular platform harnesses the same three manipulation
mechanisms: rolling, sliding, and pulling scaled across
interconnected modules. Rolling and sliding are particularly
leveraged for transferring objects between neighbouring modules,
while each module retains independent manipulation capability.
The efficiency of manipulation depends on the interplay between
object geometry and fabric properties such as friction,
elasticity, and surface tension.

\subsection{Hardware Implementation}

The physical 2$\times$2 prototype (Fig.~\ref{fig:mutli-modular MANTA-RAY}A--D)
consists of nine linear actuators connected by a scissor
mechanism that enables adjustable actuator spacing for different
workspace requirements. For the experiments in this work,
actuators are spaced 0.5~m apart, resulting in a platform size
of 1$\times$1~m with a resting height of 0.7~m. Each actuator
provides a 0.4~m vertical stroke using a stepper motor (Nema~23
Bipolar, 1.8$^\circ$ step angle, 1.85~Nm holding torque) that
converts rotational motion to linear displacement through a
pulley-belt mechanism, mounted on a linear guide for smooth
travel. Actuators are controlled by Arduino Uno
microcontrollers with CNC~V3 shields and A4988 stepper drivers,
with real-time position feedback from AS5600 magnetic encoders
(12-bit resolution). A proportional controller on each Arduino
converts desired actuator positions to motor speed commands.

The manipulation surface is a 1.2$\times$1.2~m sheet of 100\%
polyester fabric, with each module covered by a 0.6$\times$0.6~m
section that facilitates surface draping between actuator
attachment points. Object tracking is performed by a six-camera
OptiTrack system at up to 120~Hz. By modulating the relative
heights of the nine actuators, the system generates controlled
surface gradients that guide object motion across the fabric
through coordinated tilting and deformation.

%% file: method.tex
\section{Methods}

The following sections describe the simulation environment
(\ref{subsec:simulation}) and the control formulation (\ref{subsec:controller}) for the
multi-modular platform.

\subsection{Simulation Setup:}\label{subsec:simulation}
% MuJoCo \cite{todorov2012mujoco} was used as a physics-based simulator to test the controller in a simulation environment. The \textit{flexcomp} and \textit{composite} elements in MuJoCo facilitate the creation of deformable objects, allowing flexible components such as cloth or soft bodies to be defined. The simulation includes three modular platforms: a single-module setup and grids of $2 \times 2$ and $3 \times 3$ modules (4 and 9 modules, respectively). Positional actuators with a range of $[-0.20,0.20]$~m are placed at a distance of $0.5$~m from each other. A soft fabric is connected on top of the actuator grid to closely mimic the hardware system. The simulation runs at $1000$Hz. Figure\ref{fig:mutli-modular MANTA-RAY} illustrates the setup.  
MuJoCo~\cite{todorov2012mujoco} was used as a physics-based simulator, with
deformable fabric modeled using flexcomp composite elements.
The simulation includes 3$\times$3 module
configurations with positional actuators (range
$[\pm0.20]$~m) spaced 0.5~m apart and a soft fabric surface
mimicking the hardware system. The simulation runs at 1000~Hz
(Fig. \ref{fig:mutli-modular MANTA-RAY}E).

\subsection{Controller}\label{subsec:controller}

The multi-modular platform achieves object manipulation through
two core control modes. The primary challenge is inter-module
object passing, which transfers objects across module boundaries
by coordinating actuator motions on a continuously deformable
surface. Once an object reaches its target module, a local
position controller handles fine manipulation within that
module. Both rely on geometric deformation principles combined
with closed-loop feedback, maintaining a low-dimensional (2D)
control space at a control frequency of 10~Hz.

\subsubsection{Object Passing}

Objects are transferred between modules using rolling or sliding behaviors, depending on their shape, size, and weight. To induce these behaviors, actuators belonging exclusively to the current module (i.e., not shared with the neighboring module) are raised, while the remaining four actuators; two shared and two belonging to the adjacent module are lowered. This coordinated actuation creates a temporary local slope that allows the object to roll, slide, or exhibit a combination of both for smooth transfer across module boundaries.

The passing process operates hierarchically
(Algorithm~\ref{algo:multi_pass}): a high-level Manhattan-based
planner determines the module sequence, while a low-level layer
executes each transfer through timed actuator adjustments with
real-time OptiTrack feedback and adaptive retries.

% \begin{algorithm}[h]
% \caption{Object Passing Between Modules}
% \label{algo:multi_pass}
% \begin{algorithmic}[1]
%     \Require Object position $\mathbf{p}$, target position $\mathbf{p}_d$, tracker feedback
%     \Ensure Successful inter-module transfer
%     \State $m_{curr} \gets \text{DetectModule}(\mathbf{p})$, $m_{target} \gets \text{DetectModule}(\mathbf{p}_d)$
%     \While{$m_{curr} \neq m_{target}$}
%         \State path $\gets$ FindPath($m_{curr}, m_{target}$)
%         \For{each $(m_{from}, m_{to}) \in$ path}
%         \State actuators $\gets$ GetActuatorsToRaise($m_{from}, m_{to}$)
%         \State Raise exclusive actuators in $m_{from}$ opposite to $m_{to}$
%         \State Lower shared and neighboring actuators to form slope
%         \State Hold for $t_{raise}$, then settle for $t_{settle}$
%         \State Update $\mathbf{p}$, $m_{curr} \gets \text{DetectModule}(\mathbf{p})$
%         \If{$m_{curr} \neq m_{to}$}
%             \State Adapt $h_{raise}$ and $t_{raise}$ for retry
%         \EndIf
%         \EndFor
%     \EndWhile
% \State \Return success
% \end{algorithmic}
% \end{algorithm}
\begin{algorithm}[h]
\caption{Object Passing Between Modules}
\label{algo:multi_pass}
\begin{algorithmic}[1]
    \Require Object position $\mathbf{p}$, target position
    $\mathbf{p}_d$, direction $dir \in \{$center, left,
    right$\}$, tracker feedback
    \Ensure Successful inter-module transfer
    \State $m_{\text{curr}} \gets
    \text{DetectModule}(\mathbf{p})$, \; $m_{\text{target}} \gets
    \text{DetectModule}(\mathbf{p}_d)$
    \While{$m_{\text{curr}} \neq m_{\text{target}}$}
        \State path $\gets$
        \Call{FindPath}{$m_{\text{curr}}, m_{\text{target}}$}
        \For{each $(m_{\text{from}}, m_{\text{to}}) \in$ path}
            \State act $\gets$
            \Call{GetActuatorsToRaise}{$m_{\text{from}},
            m_{\text{to}}, dir$}
            \State Raise act in $m_{\text{from}}$ opposite to
            $m_{\text{to}}$
            \If{$dir =$ left}
                \State Raise inner shared actuator (toward A4)
            \ElsIf{$dir =$ right}
                \State Raise outer shared actuator (away from A4)
            \Else
                \State Raise both shared actuators equally
            \EndIf
            \State Lower remaining actuators to form slope
            \State Hold for $t_{\text{raise}}$, then settle for
            $t_{\text{settle}}$
            \State Update $\mathbf{p}$, \; $m_{\text{curr}} \gets
            \text{DetectModule}(\mathbf{p})$
            \If{$m_{\text{curr}} \neq m_{\text{to}}$}
                \State Increase $h_{\text{raise}}$ and
                $t_{\text{raise}}$ for retry
            \EndIf
        \EndFor
    \EndWhile
    \State \Return success
\end{algorithmic}
\end{algorithm}

% In Algorithm~1, $\mathbf{p}$ and $\mathbf{p}_d$ denote the
% current and target object positions, while $m_{\text{curr}}$ and
% $m_{\text{target}}$ are the corresponding modules.
% \texttt{DetectModule()} identifies the module containing the
% object, \texttt{FindPath()} computes the shortest module
% sequence via a Manhattan planner, and
% \texttt{GetActuatorsToRaise()} selects actuators to elevate for
% initiating transfer. If the object fails to reach the next
% module, $h_{\text{raise}}$ and $t_{\text{raise}}$ are adaptively
% increased for retry (lines 9--11).

In Algorithm~\ref{algo:multi_pass}, $\mathbf{p}$ and
$\mathbf{p}_d$ denote the current and target object positions,
while $m_{\text{curr}}$ and $m_{\text{target}}$ represent the
corresponding source and destination modules.
\texttt{DetectModule()} identifies the module containing the
object, and \texttt{FindPath()} computes the shortest module
sequence via a Manhattan-based planner that enforces
conflict-free paths: when multiple objects are manipulated
simultaneously, their planned module sequences are guaranteed
not to intersect, ensuring that actively manipulated objects
don't occupy same modules. This prevents
interference.
\texttt{GetActuatorsToRaise()} selects actuators to elevate
based on the transfer direction and module pair. The direction
parameter $dir$ controls the lateral passing path: raising the
inner shared actuator (toward A4) produces a left path, raising
the outer shared actuator yields a right path, and raising both
equally produces a center path, corresponding to the three
transfer trajectories validated in
Section~\ref{subsec:directional_passing}. If the object fails to
reach the next module, $h_{\text{raise}}$ and
$t_{\text{raise}}$ are adaptively increased for retry.

% In Algorithm~\ref{algo:multi_pass}, $\mathbf{p}$ and $\mathbf{p}_d$ denote the current and target object positions, respectively, while $m_{curr}$ and $m_{target}$ represent the corresponding source and destination modules. The function \texttt{DetectModule()} determines the module containing the object based on its position, and \texttt{FindPath()} computes the shortest sequence of intermediate modules using a Manhattan-based path planner. The function \texttt{GetActuatorsToRaise()} selects the actuators to be elevated for initiating the object’s movement toward the next module. Parameters $h_{raise}$ and $t_{raise}$ define the height and duration of actuator elevation, respectively, while $t_{settle}$ ensures stabilization after each transfer phase. If the object does not pass to the next module (line 10 -12) then the $h_{raise}$ and $t_{raise}$ are increased from preferred value. The process iterates until the object successfully reaches the target module, confirmed by position feedback from the tracker.

\subsubsection{Position Control}

When both the object and target lie within the same module, a local PID-based controller governs the four actuators supporting the soft surface. Each module forms a rectangular workspace of size \(m \times n\)~m, providing four degrees of freedom (DOF) for deformation. The controller takes as input the desired target position \(\mathbf{p}_d = (x_d, y_d)\), the current object position \(\mathbf{p} = (x, y)\), actuator base coordinates \(\{(x_i, y_i)\}_{i=1}^{4}\), and a reference height \(z_0 = 1.5~\mathrm{m}\), following the Manhattan controller in \cite{11163787}.

At each control step (\(f = 10~\mathrm{Hz}\)), positional errors \(e_x, e_y\) are processed by a PID controller to compute surface tilt angles \(\theta_{zx}\) and \(\theta_{zy}\) about the \(X\)- and \(Y\)-axes. The surface orientation is represented by the normal vector
\begin{equation}
    \vec{n} = (\sin\theta_{zx}, \; \sin\theta_{zy}, \; \cos\theta_{zx}),
\end{equation}
which defines the plane
\begin{equation}
    \vec{n}\cdot(\vec{r}-\vec{r_0})=0, \quad \vec{r_0}=(0,0,z_0).
\end{equation}

The desired actuator heights are then derived from the plane geometry as
\begin{equation}
    z_i = \frac{z_0 n_z - n_x x_i - n_y y_i}{n_z},
\end{equation}
with corresponding actuator commands
\begin{equation}
    a_i = (z_0 - z_i) + \alpha \delta_i, \quad \delta_i \sim U(-1,1),
\end{equation}
where the noise term \(\alpha \delta_i\) introduces small vibrations to reduce static friction and enhance object mobility. Each actuator operates within a range of \([-0.2, 0.2]~\mathrm{m}\), allowing a maximum surface tilt of approximately \(\pm38^{\circ}\). Continuous updates of tilt angles and actuator positions ensure smooth, real-time manipulation.

For multi-module operation, neighboring modules share boundary actuators that are jointly controlled to form localized gradients for object transfer. This distributed approach maintains the simplicity and low-dimensionality of the single-module controller while enabling seamless, large-scale manipulation across the deformable fabric surface.

%% file: results.tex
\section{Results}

% To evaluate the proposed platform, we conducted
% five sets of experiments on the 2$\times$2 hardware prototype,
% each targeting a distinct aspect of system performance:
% (1)~\textit{Object Passing} characterises the trajectory and
% repeatability of inter-module transfer for objects with diverse
% geometries; (2)~\textit{Object Behavior} examines how transfer
% position relative to shared actuators affects manipulation paths;
% (3)~\textit{Target Reaching} evaluates positioning accuracy of
% the controller across modules for heterogeneous objects;
% (4)~\textit{Multi-Object Manipulation} demonstrates parallel,
% decentralised manipulation of multiple objects simultaneously;
% and (5)~\textit{Influence on Neighbouring Module} quantifies the
% interference caused by shared boundary actuators during
% simultaneous operation and evaluates the proposed stabilisation
% strategies. We validate scalability through MuJoCo
% simulations on 3$\times$3 module configurations.
% The following subsections describe each experiment and its
% results in detail.

To evaluate the proposed platform, we conducted five sets of
experiments on the 2$\times$2 hardware prototype. The first
three experiments, ~\textit{Object Passing}, ~\textit{Directional Passing}, and
~\textit{Target Reaching}, collectively demonstrate reliable inter-module
object transfer on a soft deformable surface for the first time,
characterising transfer repeatability, directional control, and
cross-module positioning accuracy. The remaining two
experiments, ~\textit{Multi-Object Manipulation} and ~\textit{Influence on Neighbouring Module}, evaluate the consequences of
shared-boundary actuation for parallel operation. We validate
scalability through MuJoCo simulations on 3$\times$3 module
configurations.

% To evaluate the effectiveness of Modular MANTA-RAY system, we conducted three types of experiments on the actual hardware setup:

% 	1.	Object Passing: This experiment was conducted to determine how the relative elevation of the actuators affects the dynamic path of the object while passing the object to neighboring module, depending on its shape.
    
% 	2.	Object Behavior: In this experiment, various objects were passed between all modules in square trajectory on the platform. This information is crucial to identify potential extent and limitations of manipulation.
	
%     3.	Target Reaching: The controller was used to move an object from its initial position to a specified goal location on the platform. This experiment was designed to evaluate the effectiveness and robustness of the controller when managing the non-linear behavior of the soft surface.
    
%     4. Multi-object Handling: Multiple objects were place on the modules and directed to their respective target positions to check the multi-object handling capabilities of the platform.

% Following subsection describe the experimental setup and results of these experiments in more details. 

\subsection{Object passing}
\begin{figure*}[!t]
    % \centering
    % First row with three columns
    \begin{subfigure}[b]{0.31\textwidth}
        \centering
        \includegraphics[width=\textwidth]{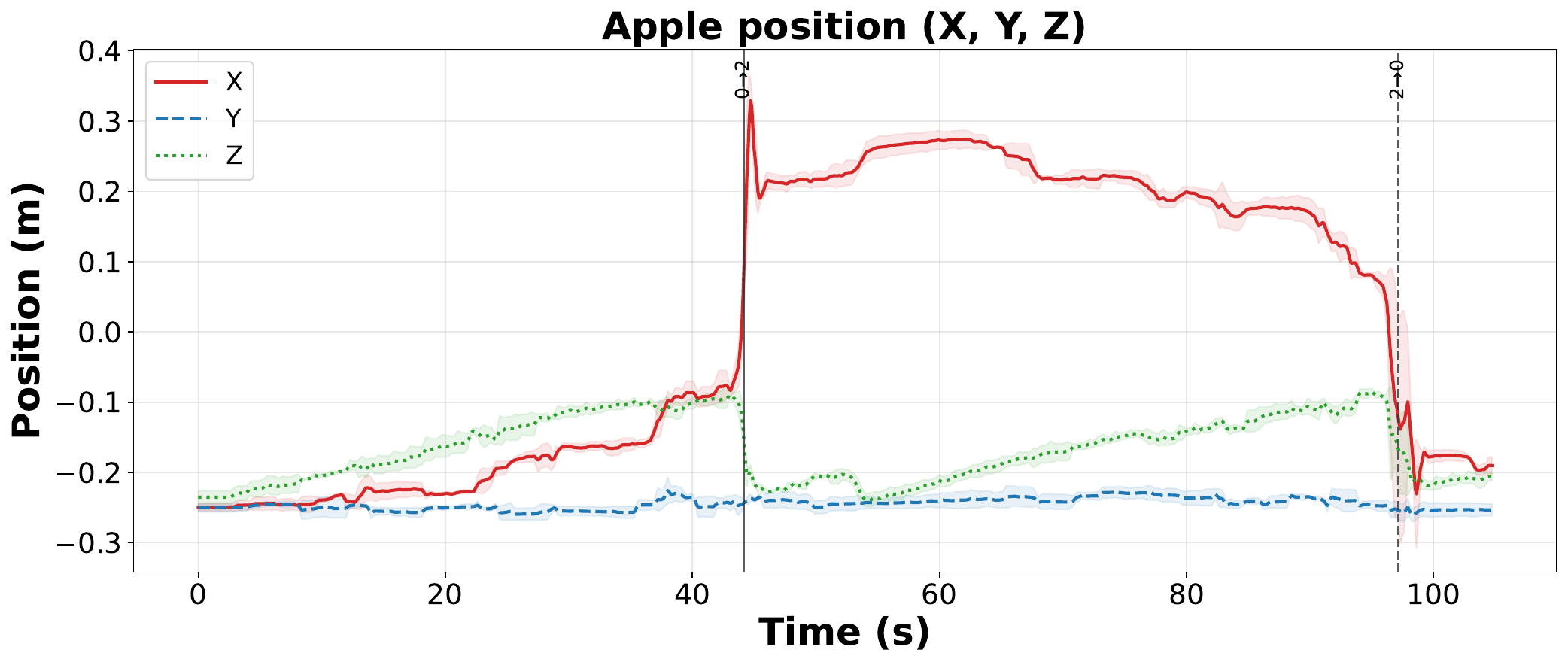} % Replace with your image path
        \caption{Apple}
        \label{fig:p_apple}
    \end{subfigure}
    \hfill
    \begin{subfigure}[b]{0.31\textwidth}
        \centering
        \includegraphics[width=\textwidth]{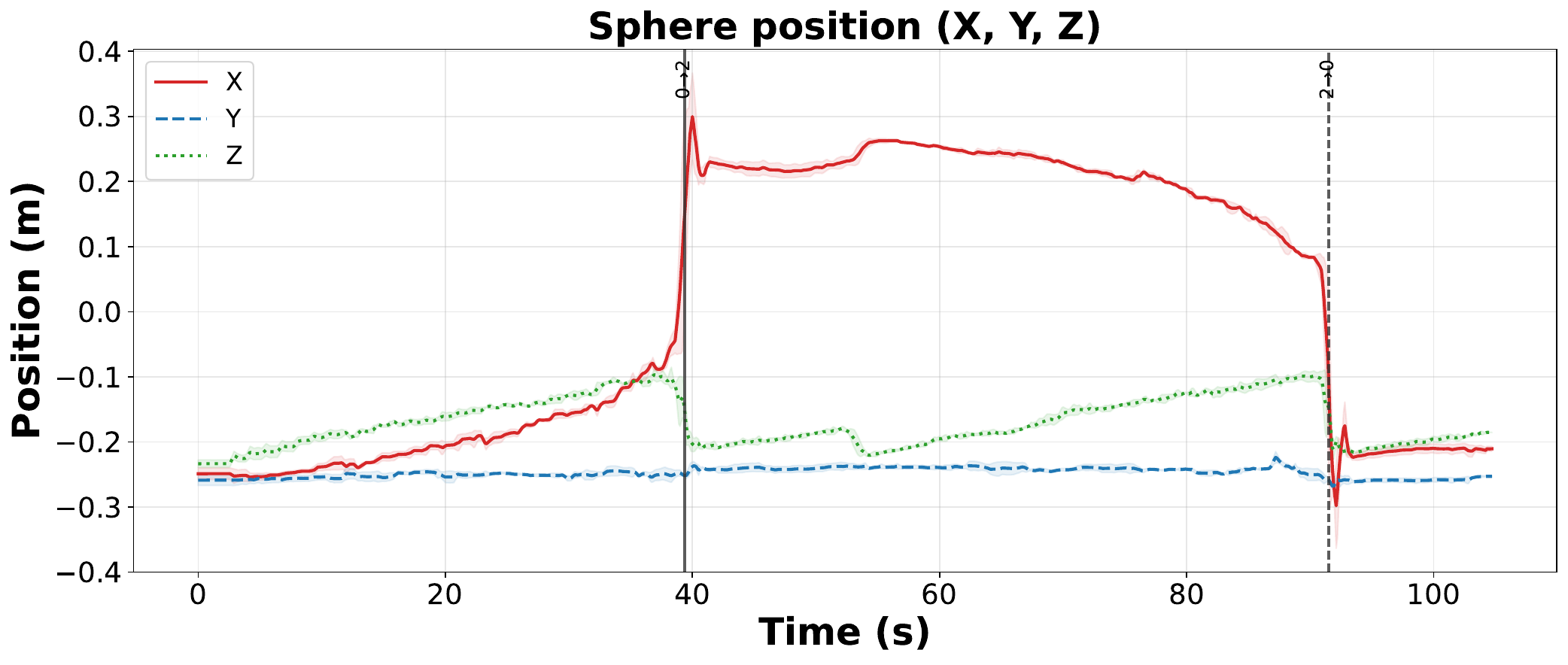} % Replace with your image path
        \caption{Sphere}
        \label{fig:p_sphere}
    \end{subfigure}
    \hfill
    \begin{subfigure}[b]{0.31\textwidth}
        \centering
        \includegraphics[width=\textwidth]{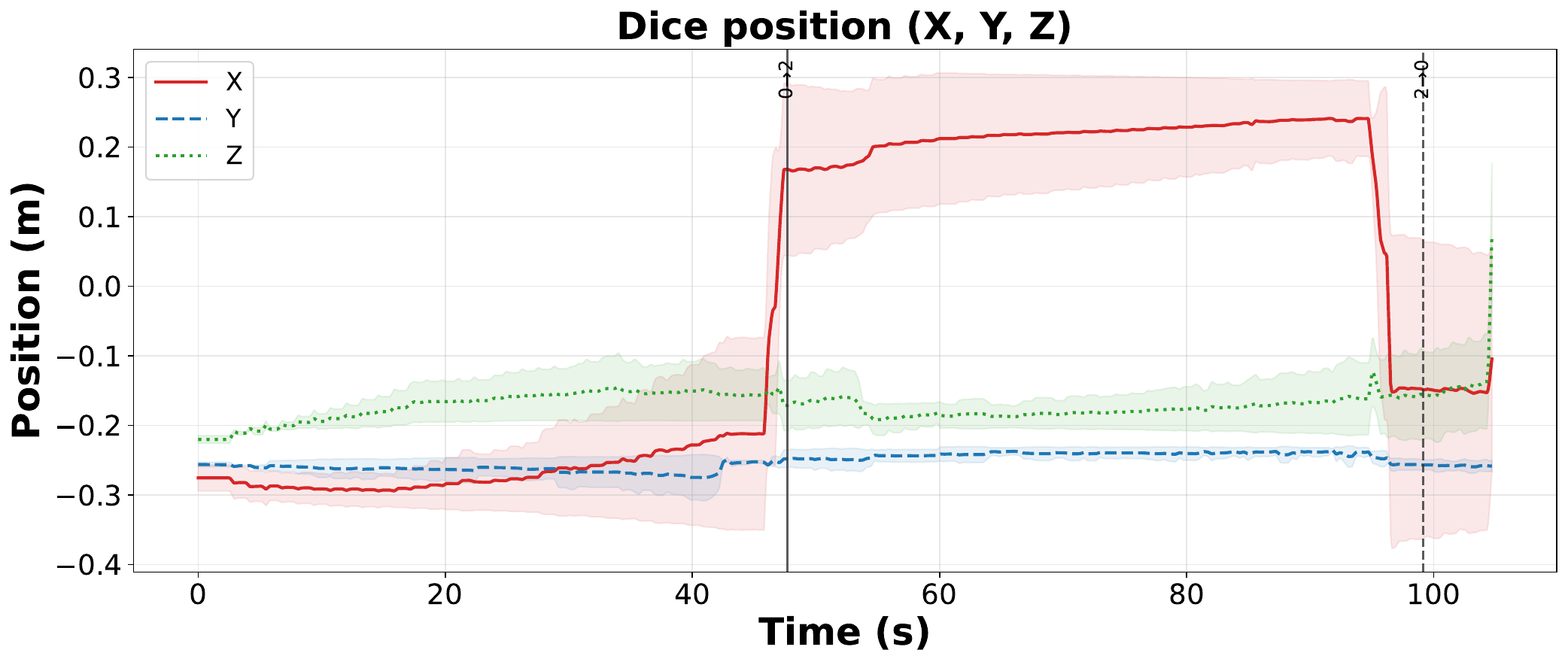} % Replace with your image path
        \caption{Dice}
        \label{fig:p_die}
    \end{subfigure}
    \hfill
    \begin{subfigure}[b]{0.31\textwidth}
        \centering
        \includegraphics[width=\textwidth]{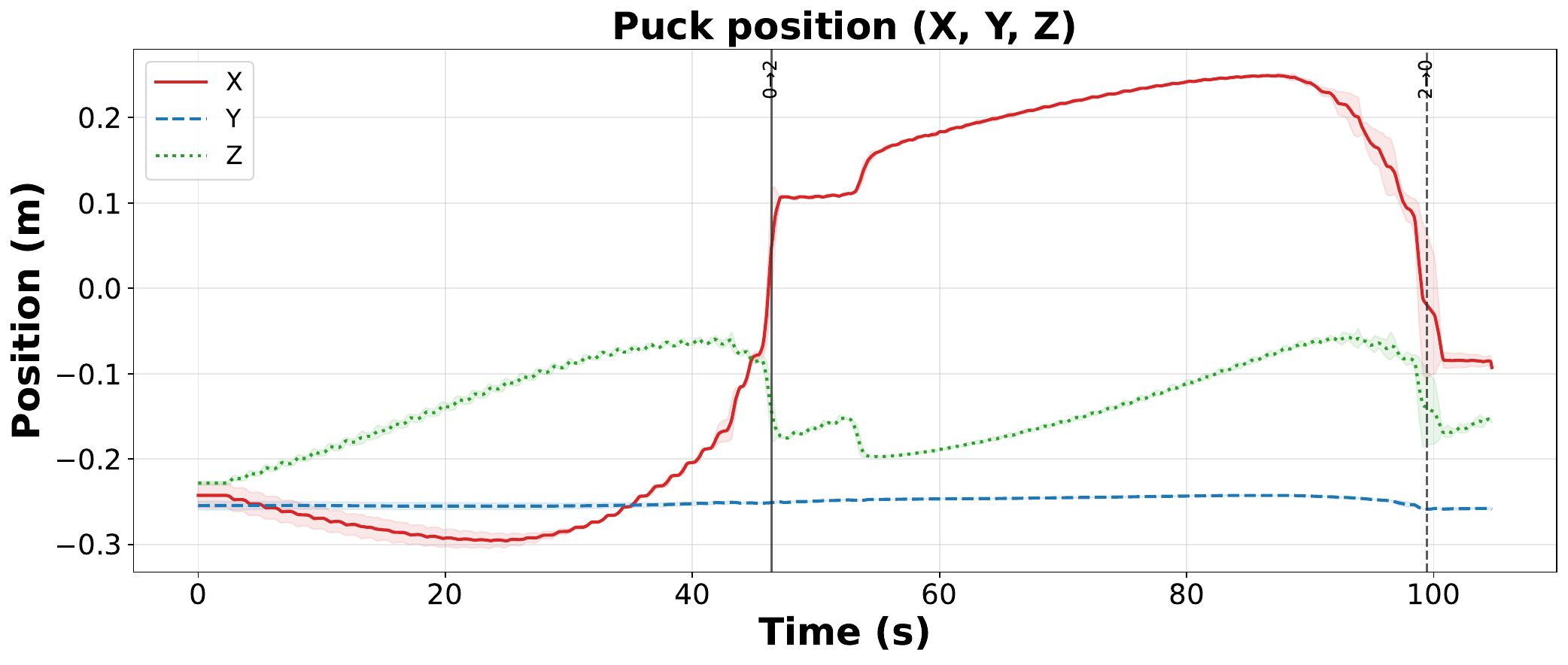} % Replace with your image path
        \caption{Puck}
        \label{fig:p_puck}
    \end{subfigure}
    \hfill
    \begin{subfigure}[b]{0.31\textwidth}
        \centering
        \includegraphics[width=\textwidth]{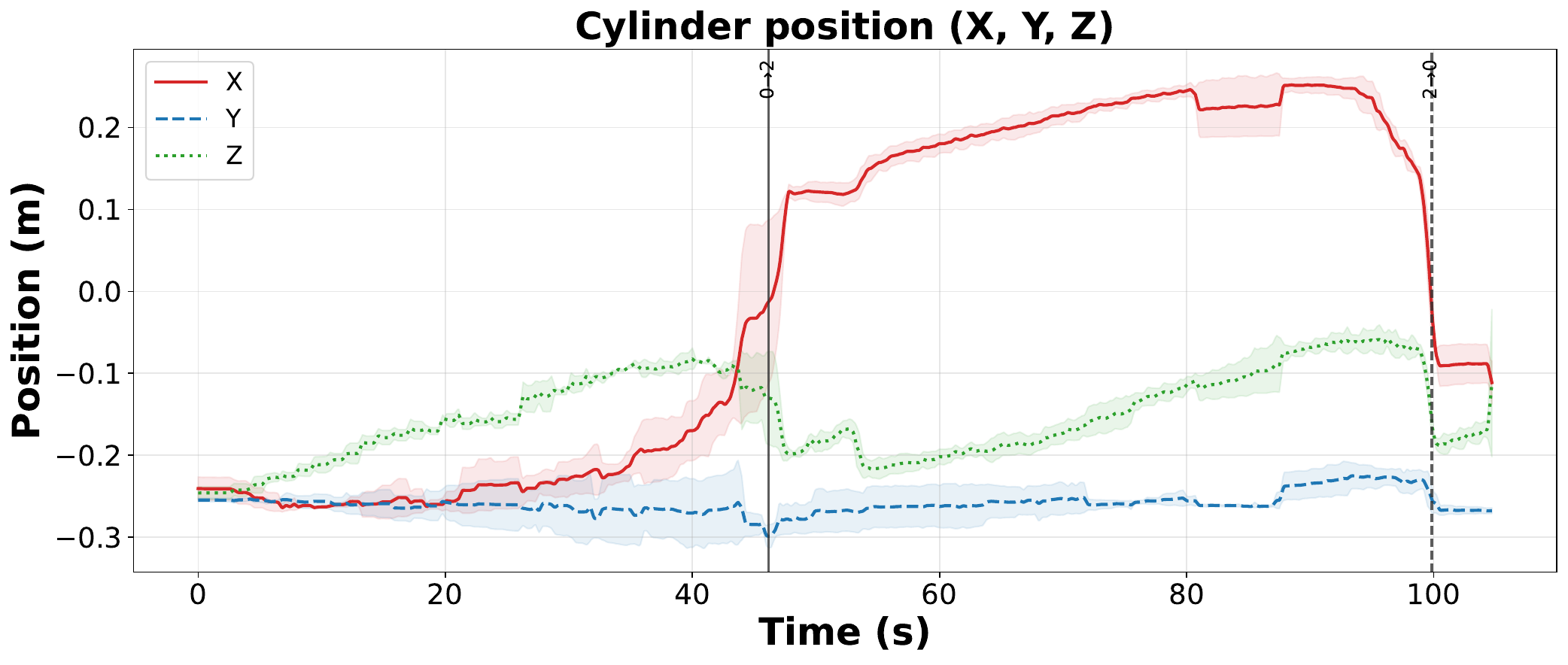} % Replace with your image path
        \caption{Cylinder}
        \label{fig:p_cylinder}
    \end{subfigure}
    \hfill
    \begin{subfigure}[b]{0.31\textwidth}
        \centering
        \includegraphics[width=\textwidth]{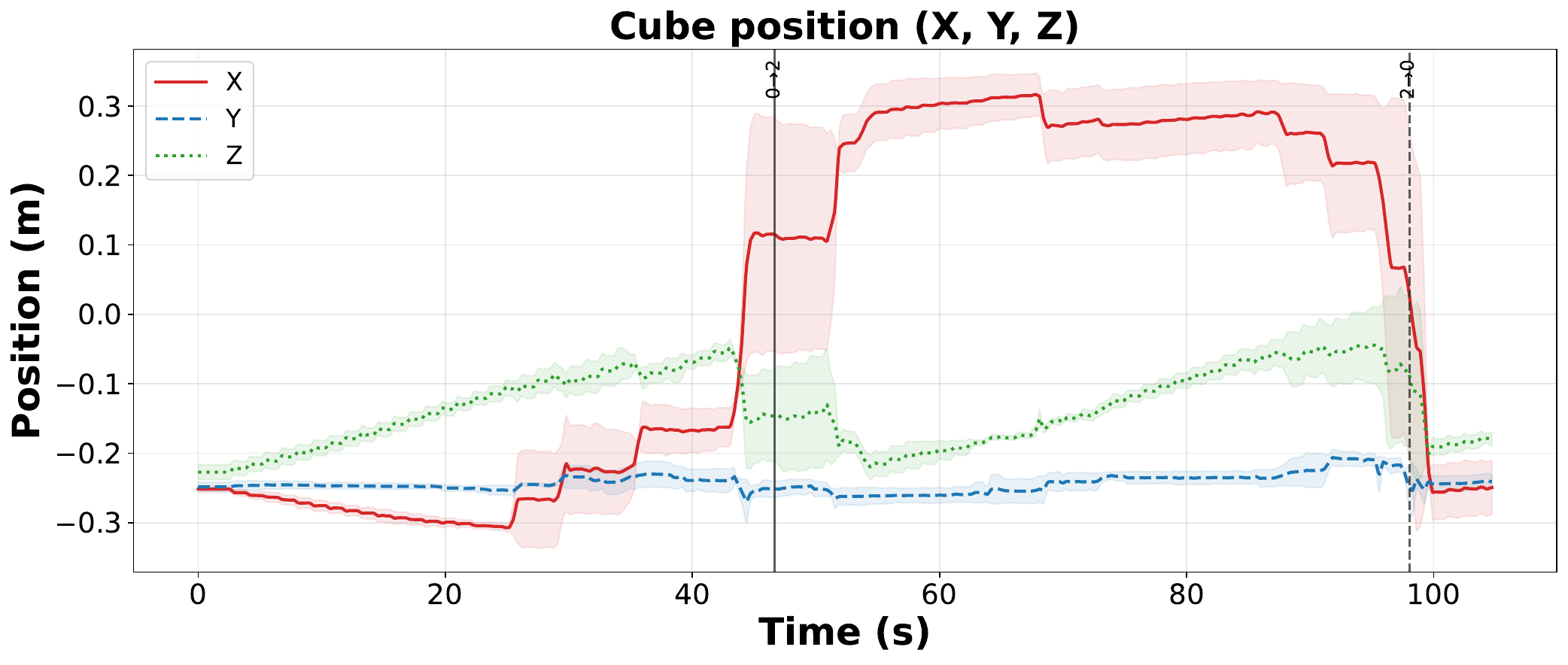} % Replace with your image path
        \caption{Cube}
        \label{fig:p_cube}
    \end{subfigure}
    
    \caption{Object Passing from M0 to M2 and back from M2 to M0}
    \label{fig:behaviour}
\end{figure*}

% \begin{figure}[!t]
%     \centering
%     % First row with three columns
%     \begin{subfigure}{0.5\textwidth}
%         \centering
%         \includegraphics[width=\textwidth]{figs/passing/multi_object_std_barplot_comparison.png} 
%         % \caption{Side view of the platform without the flexible surface}
%     \end{subfigure}
%     \caption{Mean Standard deviation comparison across objects during passing between modules}
%     \label{fig:p_std}
% \end{figure}

During object passing, the rolling and sliding behaviors depend strongly on the object’s shape and surface characteristics. The more irregular the object, the more distinct and variable its motion becomes during transfer. To evaluate transfer repeatability, each of six objects
(Table~\ref{tab:objects}) was passed from the center of module
M0 to adjacent module M2 and back, repeated three times.
Figure~\ref{fig:behaviour} presents the resulting trajectories,
with vertical lines indicating module boundary crossings.
% To ensure effective control, maintaining consistency in these behaviors is essential. To evaluate this, each object was passed from the center of the starting module (M0) to the center of an adjacent module (M2) and then back to M0. The experiment was repeated three times to assess behavioral repeatability across multiple trials.  

Spherical objects such as the apple (Fig.~\ref{fig:p_apple})
and sphere (Fig.~\ref{fig:p_sphere}) exhibited consistent
rolling behavior, while the disk showed repeatable sliding due
to its large contact area and low center of gravity. In
contrast, the cube (Fig.~\ref{fig:p_cube}) and dice
(Fig.~\ref{fig:p_die}) displayed greater variability across
runs: the cube rolled in multiple discrete steps, whereas the
dice tended to roll in single steps, reflecting differences in size and weight. The cylindrical object exhibited both rolling and sliding depending on its orientation, with greater variability when misaligned.  Overall, smoother symmetric objects
produced stable repeatable trajectories, while irregular
geometries increased variability in both lateral and vertical
motion. The mean positional deviation across all objects
remained within $0.03$~m, confirming consistent inter-module
transfer performance.

\subsection{Directional Passing}\label{subsec:directional_passing}

\begin{figure*}
    % \centering
    % First row with three columns
    \begin{subfigure}[b]{0.3\textwidth}
        \centering
        \includegraphics[width=\textwidth]{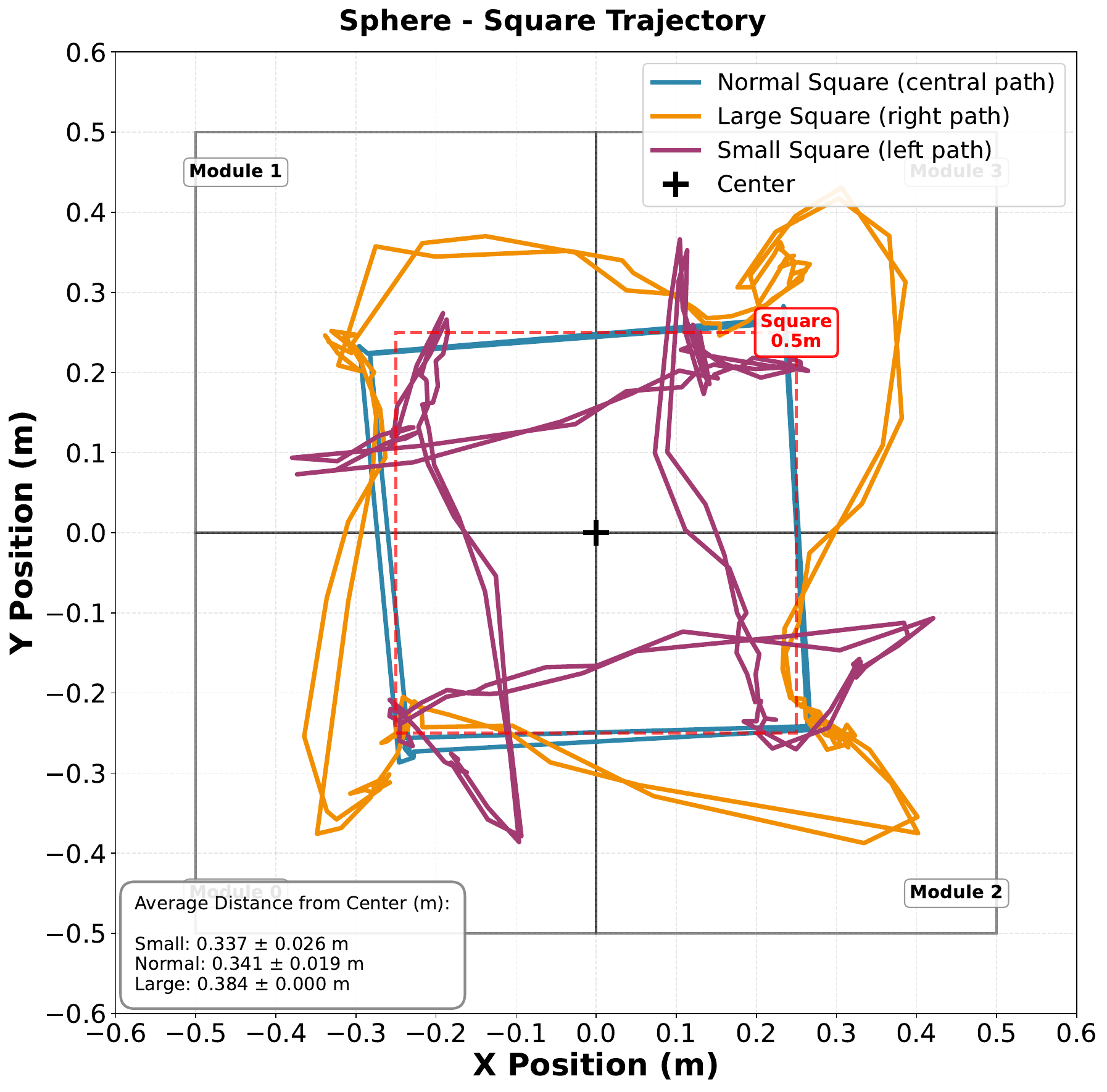} % Replace with your image path
        \caption{Sphere}
        \label{fig:subfig1}
    \end{subfigure}
    \hfill
    \begin{subfigure}[b]{0.3\textwidth}
        \centering
        \includegraphics[width=\textwidth]{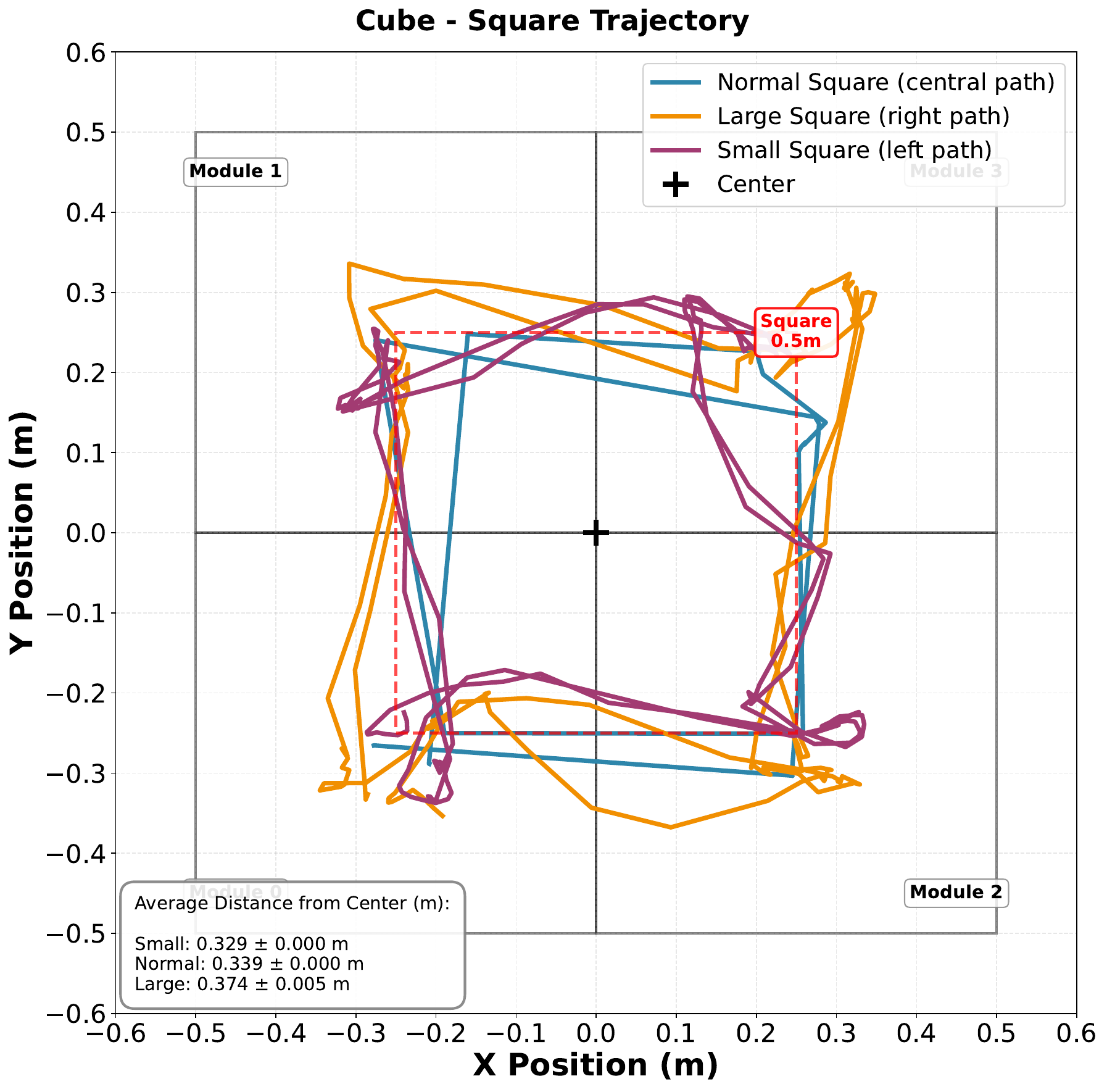} % Replace with your image path
        \caption{Cube}
        \label{fig:subfig2}
    \end{subfigure}
    \hfill
    \begin{subfigure}[b]{0.3\textwidth}
        \centering
        \includegraphics[width=\textwidth]{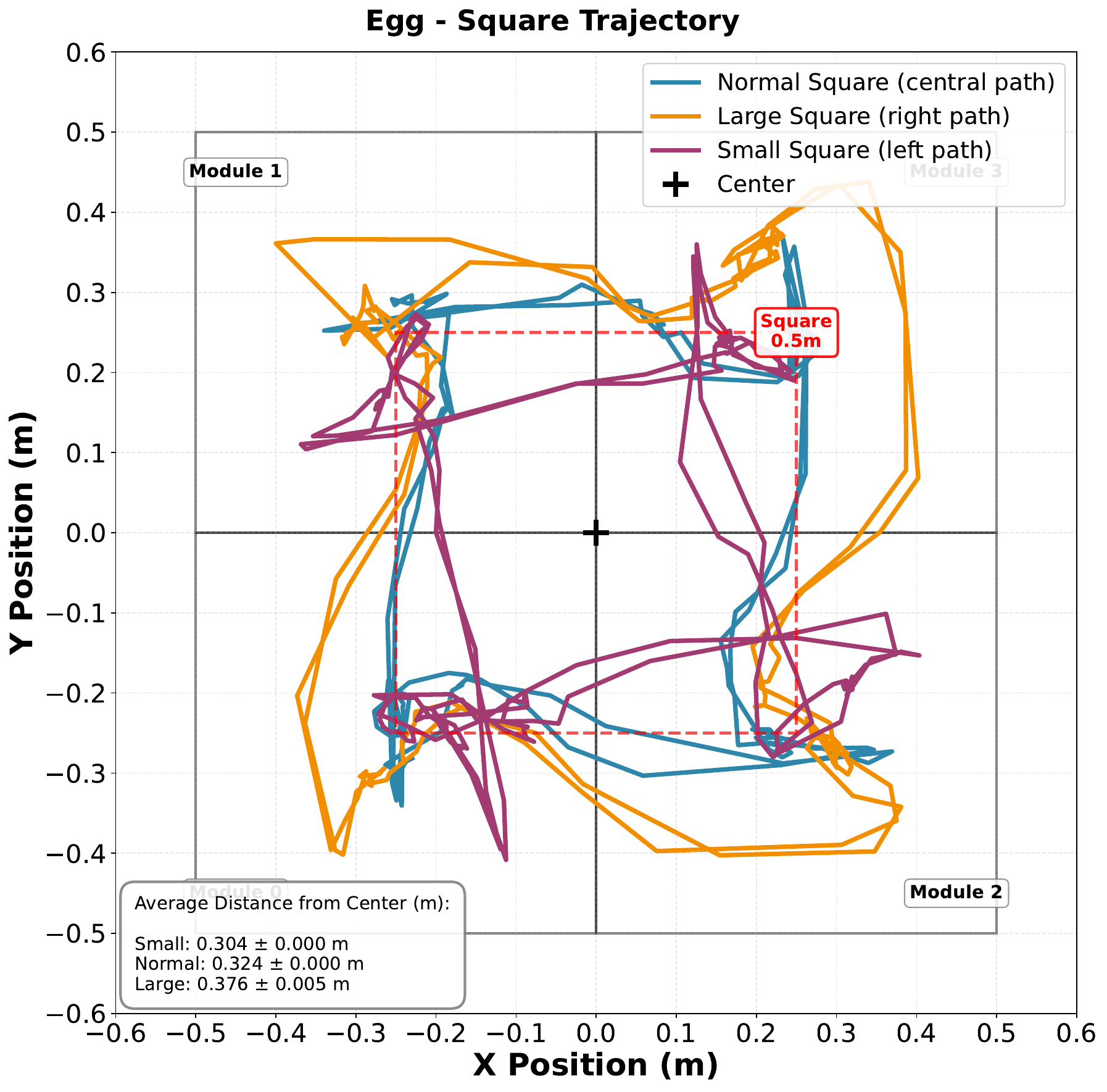} % Replace with your image path
        \caption{Egg}
        \label{fig:subfig3}
    \end{subfigure}
    
    \caption{Three square paths (anticlockwise: M0$\rightarrow$M2$\rightarrow$M3$\rightarrow$M1) of the object: Normal square (central direction, Blue), Small square ( left direction, Purple) and Large square (right direction, Orange)}
    \label{fig:sqaure_pasing}
\end{figure*}

We investigated how the lateral position of object transfer, whether along the center, left, or right path relative to
the passing direction affects manipulation performance. In
this experiment, various objects were passed between
neighbouring modules along a square trajectory
(anticlockwise: M0$\rightarrow$M2$\rightarrow$M3$\rightarrow$M1)
across the surface. Each object followed three square paths in
a continuous loop of two cycles: a \textit{normal square}
(blue), where the object passed through the center between
shared actuators; a \textit{small square} (purple), where the
transfer occurred closer to the central actuator (A4); and a
\textit{large square} (yellow), where the object was passed
farther from the central actuator (A4). Ideally, during object
passing an object should move along the central path between
the shared actuators, forming a perfect square trajectory as
shown by the dotted red line in Fig.~\ref{fig:sqaure_pasing},
representing the normal square. When the central actuator is
raised during object passing, the object traces a larger square
path, whereas raising the outer actuator between shared modules
results in a smaller square path. In practice, the trajectories
deviate slightly for normal squares due to the fabric's
elasticity and tension distribution. Spherical objects such as
the sphere and egg produced three clearly distinguishable paths
corresponding to the three square sizes, while the cube
primarily followed the central trajectory between shared
actuators. These results demonstrate that the passing direction
can be actively controlled by selecting which boundary actuator
to raise, enabling center, left, or right transfer paths
relative to the shared actuator pair.

\subsection{Target reaching}
\begin{figure}[!t]
    % \centering
    % First row with three columns
    % \begin{subfigure}[b]{0.3\textwidth}
    %     \centering
    %     \includegraphics[width=\textwidth]{figs/target/sphere_target_reaching_20251027-170124_trajectory.png} % Replace with your image path
    %     \caption{Sphere}
    %     \label{fig:t_sphere}
    % \end{subfigure}
    % \hfill
    % \begin{subfigure}[b]{0.3\textwidth}
    %     \centering
    %     \includegraphics[width=\textwidth]{figs/target/cube_target_reaching_20251027-170438_trajectory.png} % Replace with your image path
    %     \caption{Cube}
    %     \label{fig:t_cube}
    % \end{subfigure}
    % \hfill
    \begin{subfigure}[b]{0.24\textwidth}
        \centering
        \includegraphics[width=\textwidth]{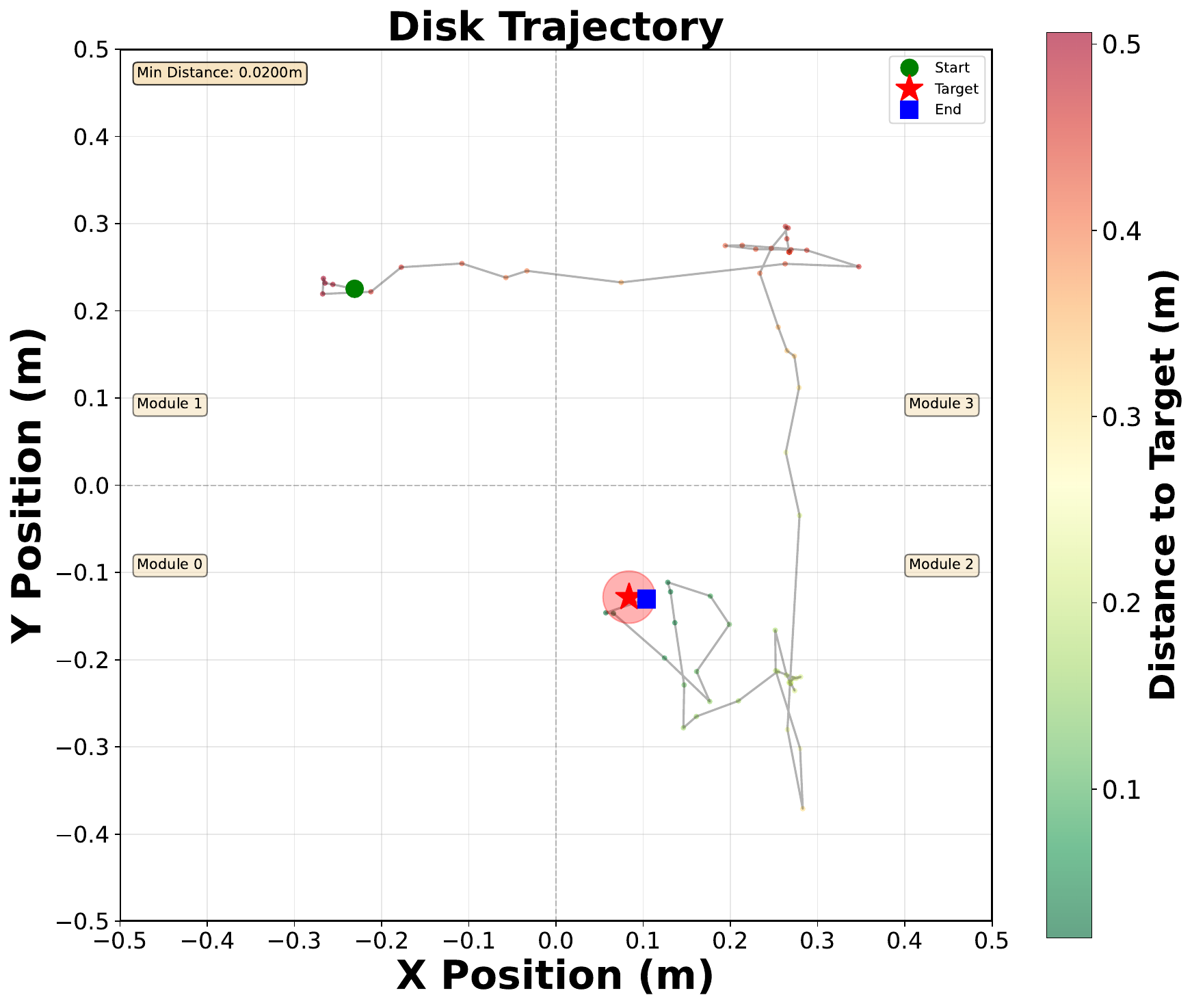} % Replace with your image path
        \caption{Puck}
        \label{fig:t_puck}
    \end{subfigure}
    \hfill
    \begin{subfigure}[b]{0.24\textwidth}
        \centering
        \includegraphics[width=\textwidth]{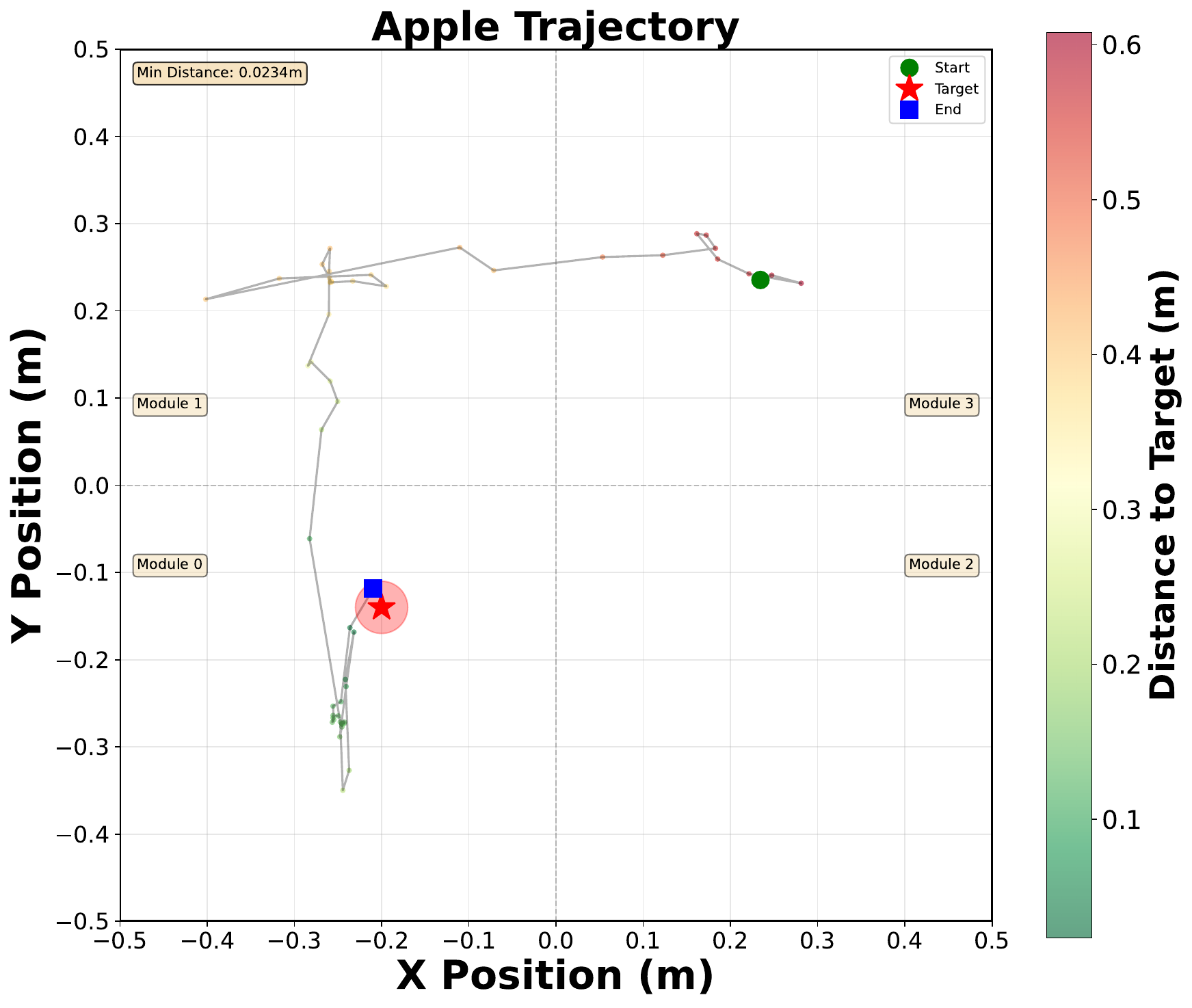} % Replace with your image path
        \caption{Apple}
        \label{fig:t_apple}
    \end{subfigure}
    \hfill
    \begin{subfigure}[b]{0.24\textwidth}
        \centering
        \includegraphics[width=\textwidth]{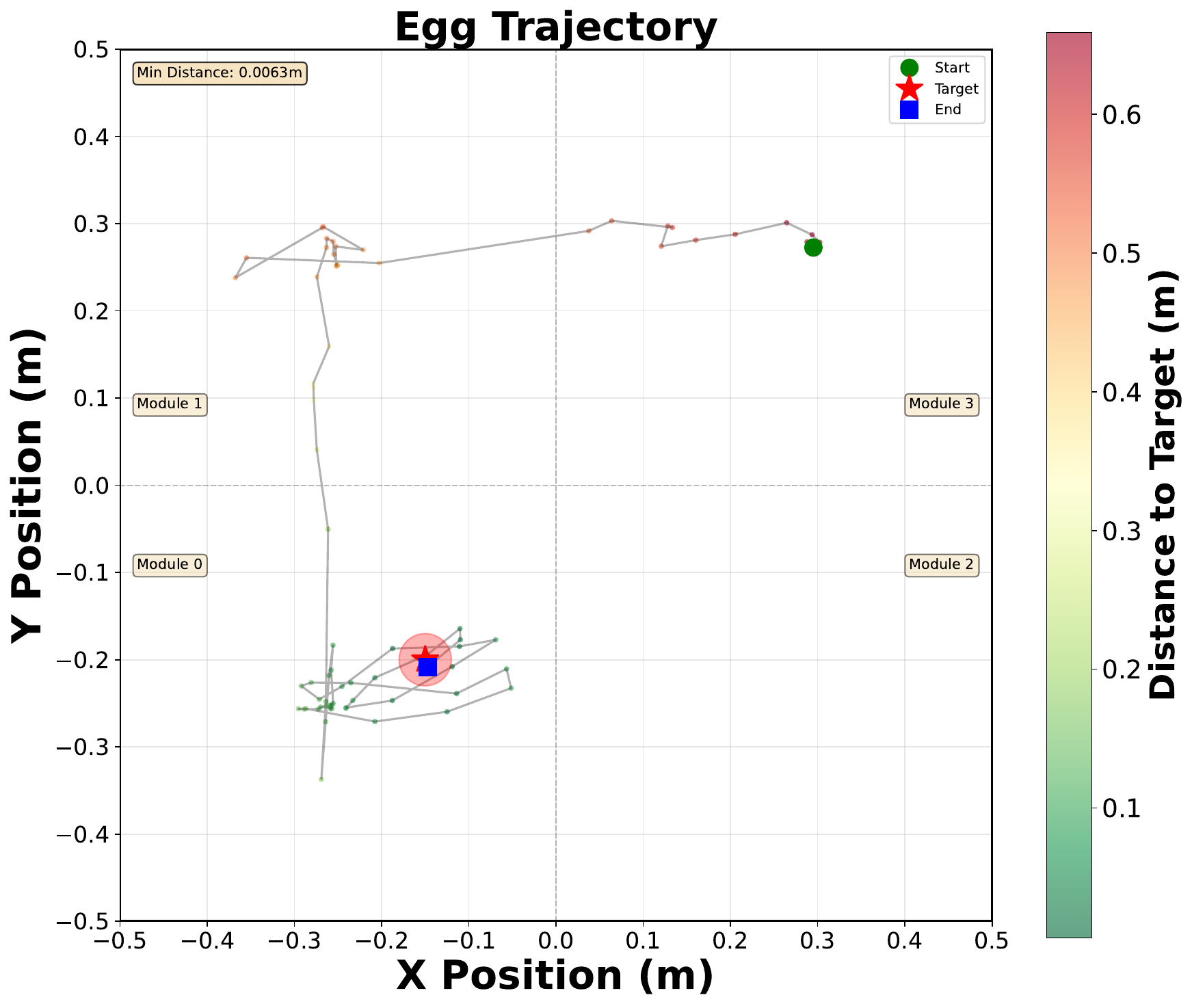} % Replace with your image path
        \caption{Egg}
        \label{fig:t_egg}
    \end{subfigure}
    \hfill
    \begin{subfigure}[b]{0.24\textwidth}
        \centering
        \includegraphics[width=\textwidth]{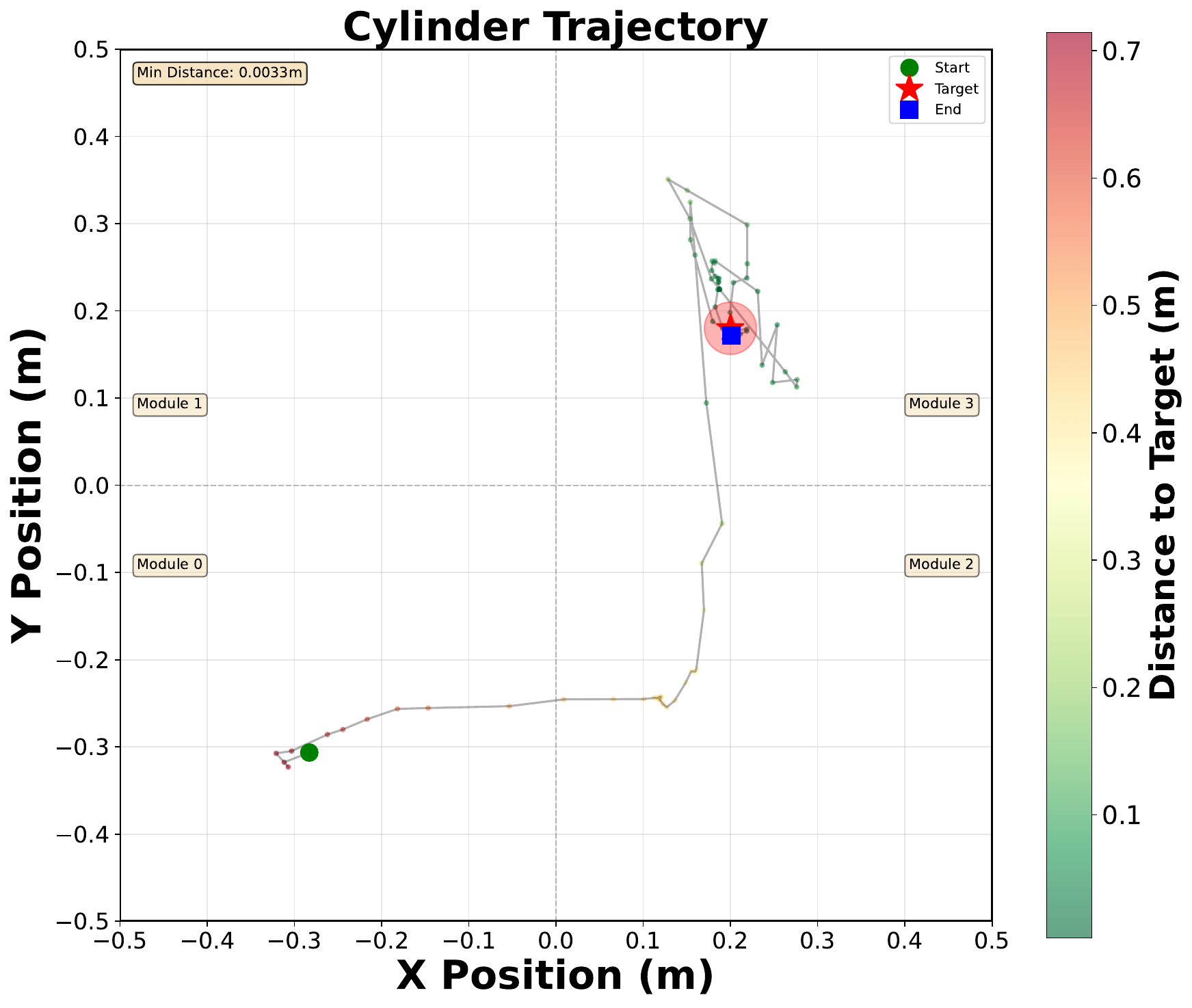} % Replace with your image path
        \caption{Cylinder}
        \label{fig:t_cylinder}
    \end{subfigure}
    
    \caption{Target reaching for different objects on hardware platform. Object start at initial position (green circle) and move to target location (red star) within a threshold of 3 cm (red circle) along the line path.} 
    \label{fig:target}
\end{figure}

Target reaching evaluates end-to-end manipulation performance,
combining inter-module transfer with local position control. We performed a target-reaching experiment with seven heterogeneous objects listed in Table~\ref{tab:objects} to evaluate the precision of manipulation; four representative cases are
shown in Fig.~\ref{fig:target}. Each object was placed at an initial position (green circle) and tasked to reach a target location (red star) within a threshold radius of 0.03~m (red circle). If the object’s initial module differed from the target module, it was first transferred along a Manhattan path to the appropriate module. Once on the target module, the position controller was activated to achieve precise alignment within the specified boundary. 

Figure~\ref{fig:target} shows that all objects successfully reached their respective target positions within the defined threshold on $2\times2$ hardware module. Elongated objects, such as the egg (Fig.~\ref{fig:t_egg}) and the cylinder (Fig.~\ref{fig:t_cylinder}), reached the target but required small corrective zig-zag motions to fine-tune their final positions. In contrast, spherical objects achieved smoother and more direct trajectories, demonstrating higher positional stability and control precision. Overall, the system achieved a mean positioning error of less than 0.02~m across all trials, confirming the controller’s reliability in handling heterogeneous objects.

\subsection{Multi-Object Manipulation}
\begin{figure}[!t]
    \centering
    % First row with three columns
    \begin{subfigure}{0.239\textwidth}
        \centering
        \includegraphics[width=\textwidth]{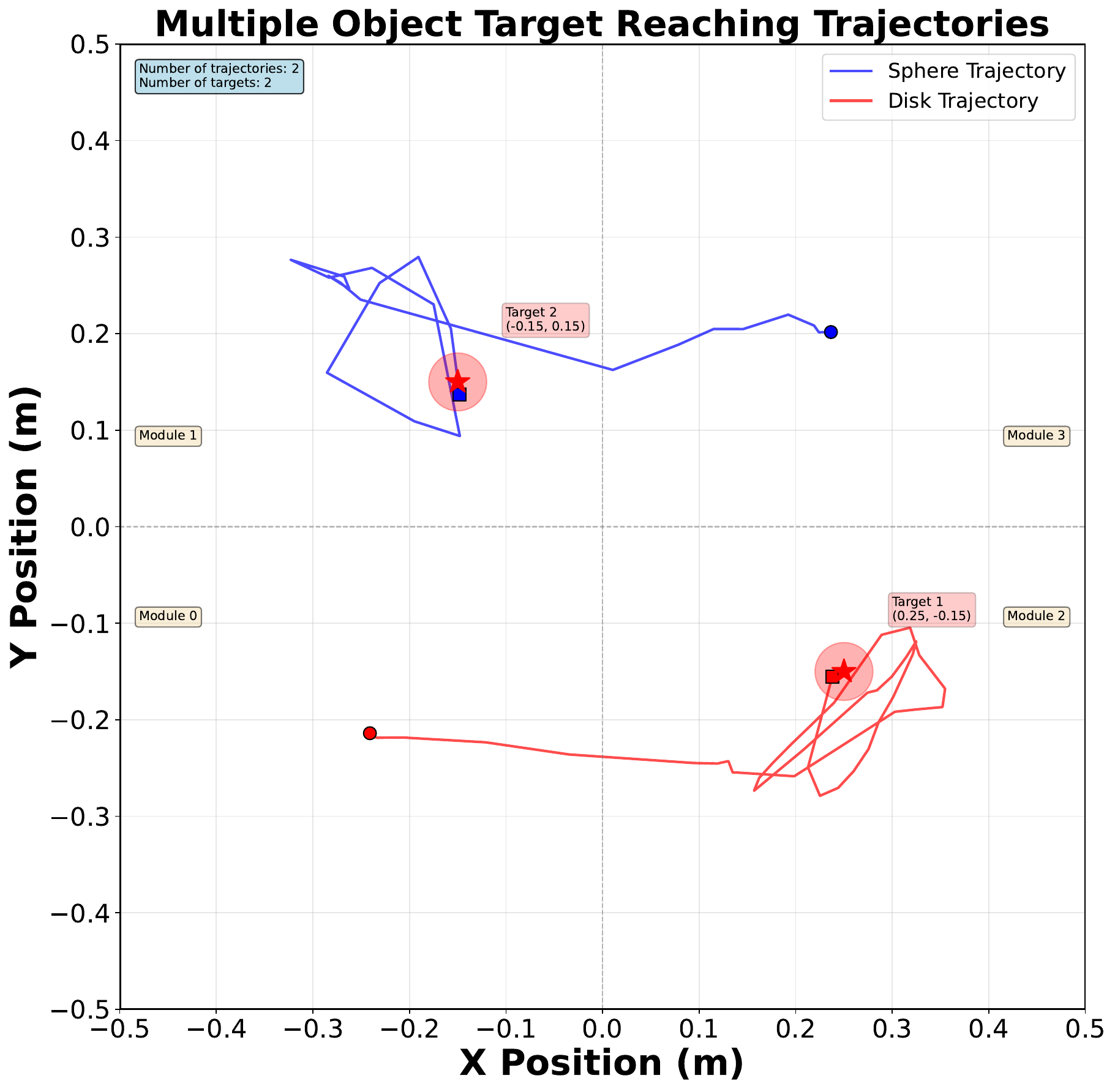} 
        \caption{$2 \times 2$ Hardware platform }
        \label{fig:mo_hardware}
    \end{subfigure}
    \hfill
    \begin{subfigure}{0.239\textwidth}
        \centering
        \includegraphics[width=\textwidth]{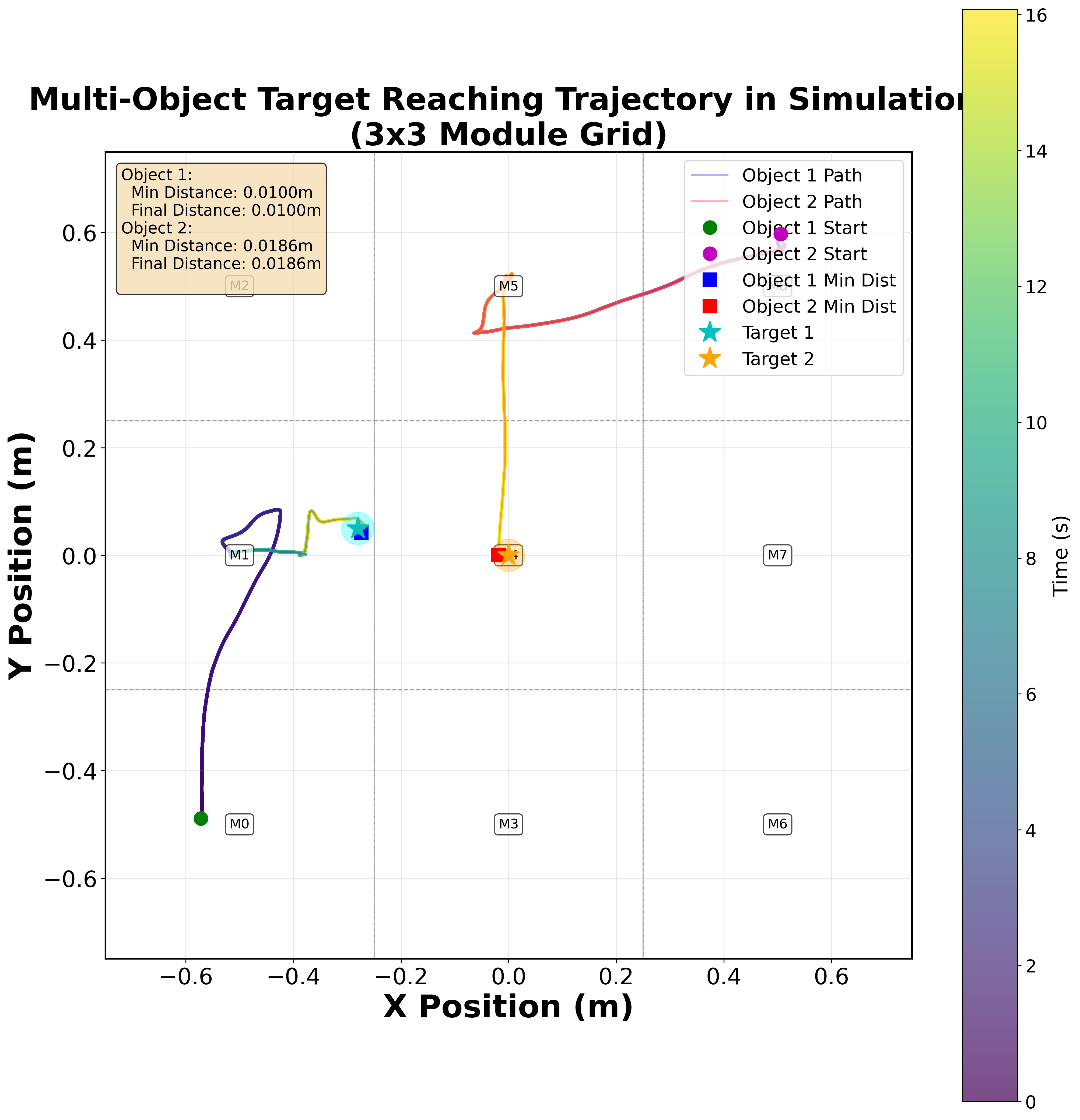} 
        \caption{$3 \times 3$ Simulation platform}
        \label{fig:mo_sim_3x3}
    \end{subfigure}
    % \hfill
    % \begin{subfigure}{0.3\textwidth}
    %     \centering
    %     \includegraphics[width=\textwidth]{figs/target/sim3x3.png} 
    %     \caption{$2 \times 2$ Simulation }
    %     \label{fig:sim_2x2}
    % \end{subfigure}
    \caption{Multi-object target reaching demonstrated on the multi-module platform: (a) hardware implementation, (b) simulation on a $3 \times 3$ modular configuration.}
    \label{fig:parallel}
\end{figure}

To evaluate the scalability and distributed coordination capability of the multi-modular platform, we conducted experiments involving simultaneous manipulation of two objects, a disk and a sphere on the $2 \times 2$ modular hardware platform. In this setup, each module operated its local controller independently, allowing parallel actuation and position control without centralized coordination. Objects were placed on different modules, a sphere on M3 and Disk on M0 and tasked to reach individual target locations M1 and M3 respectively.

Figure~\ref{fig:parallel}a shows the trajectories of a
sphere and a disk manipulated concurrently across separate
modules. The results demonstrate that the distributed control
scheme enables stable and synchronized motion across modules.
The platform maintained consistent tracking performance and each
object reached its target within the 0.03~m positional
threshold. The conflict-free Manhattan path planner ensures that
the two objects never occupy same modules
simultaneously, avoiding interference. We also
verified scalability of the platform in simulation by performing
multi-object target reaching on a 3$\times$3 platform with two
spherical objects (Fig.~\ref{fig:parallel}b). These results
confirm that the modular actuation framework supports parallel
manipulation of heterogeneous objects, achieving sub-3~cm
positioning accuracy for both objects simultaneously. 
% Overall, these results validate the scalability of the distributed multi-module MANTA-RAY control strategy. The system is capable of performing multiple independent manipulation tasks in real time while preserving the soft, compliant nature of the deformable surface, laying the groundwork for large-scale, coordinated manipulation systems.

\begin{table}[ht ]
\centering
\begin{tabular}{l l l l}
\Xhline{3\arrayrulewidth} % Thicker top line
\addlinespace[2pt] % Add space between top line and header
\textbf{Object} & \textbf{Weight (g)} & \textbf{Size (cm)} & \textbf{Fabrication Method} \\ 
\addlinespace[2pt]
\Xhline{1\arrayrulewidth} % Regular line between header and content
\addlinespace[2pt]
Sphere    & 32  & \diameter4.5           & Molding        \\ 
Cube      & 31    & 4.2 x 4.2 x 4.2               & FDM 3D Printing    \\ 
Disk      & 26   & \diameter7, Thickness: 2.5 & FDM 3D Printing    \\ 
Apple     & 120.4   & \diameter5--7        &  -- \\ 
Cylinder  & 71.3  & \diameter3, Height: 13   & Clean Spray hand sanitizer        \\ 
Egg       & 61.2  & \diameter(4.1--6.2)         &  --      \\ 
Dice      & 5    & 1.5 x 1.5 x 1.5               & Molding    \\
% Bunny     & 50.3  &  Height: 9, length: 9      & 3D Printing \\
\addlinespace[2pt] % Add space between the last row and bottom line
\Xhline{3\arrayrulewidth} % Thicker bottom line
\end{tabular}
\caption{Details of objects: weight, size, and fabrication method.}
\label{tab:objects}
\end{table}

\subsection{Influence on Neighbouring Module Object}

\begin{figure}
    \centering
    \includegraphics[width=0.8\linewidth]{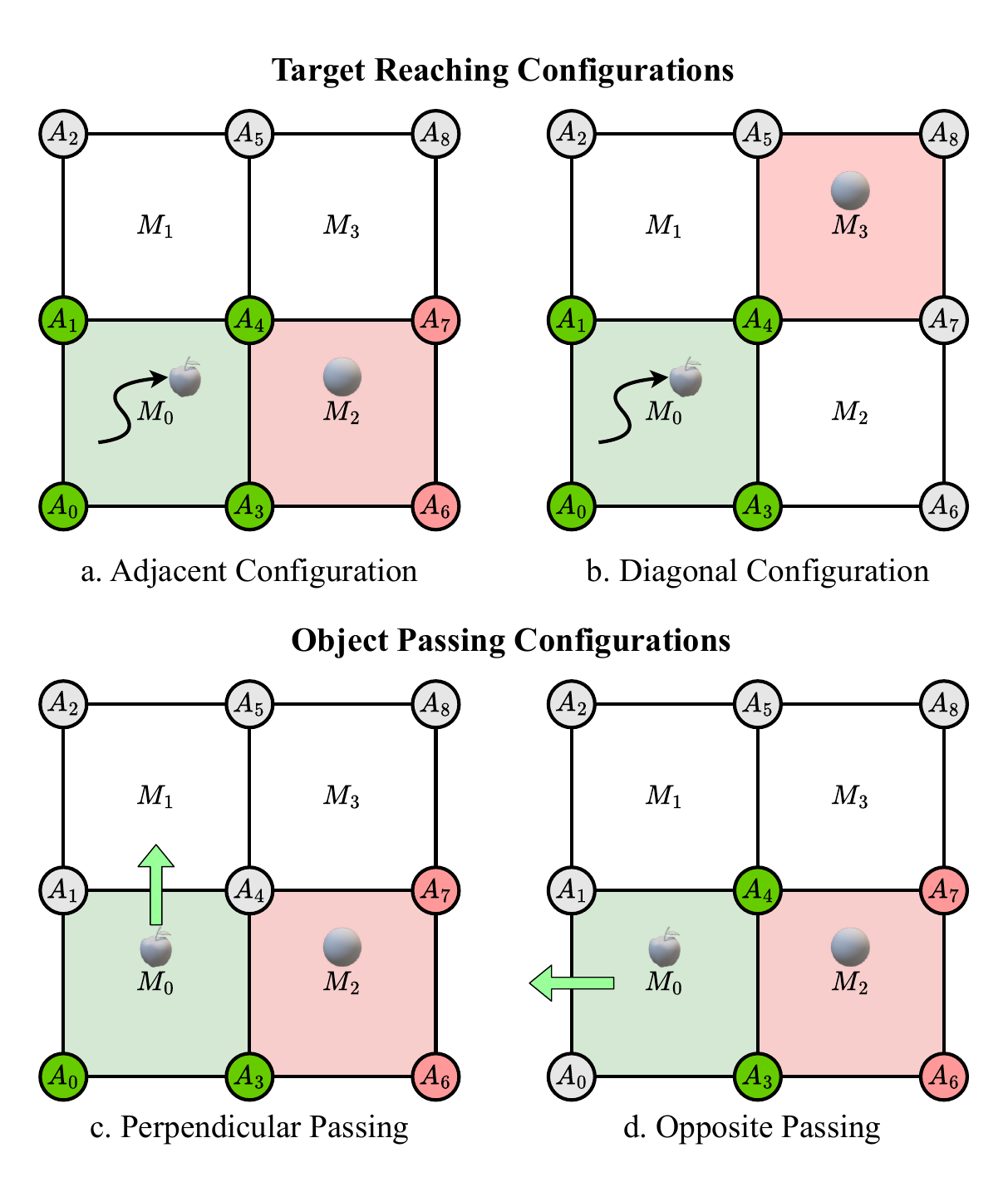}
    \caption{Spatial configurations for the neighbouring-module
    interference study ($2\times2$ Module, top view). The manipulated object
    resides on the active (green) module; the stationary object
    is tracked on the passive (red) module. For stabilisation in
    c, red actuators are raised to compensate for
    shared-actuator coupling.}
    \label{fig:interference_schematic}
\end{figure}

\begin{figure}
    \centering
    \includegraphics[width=\linewidth]{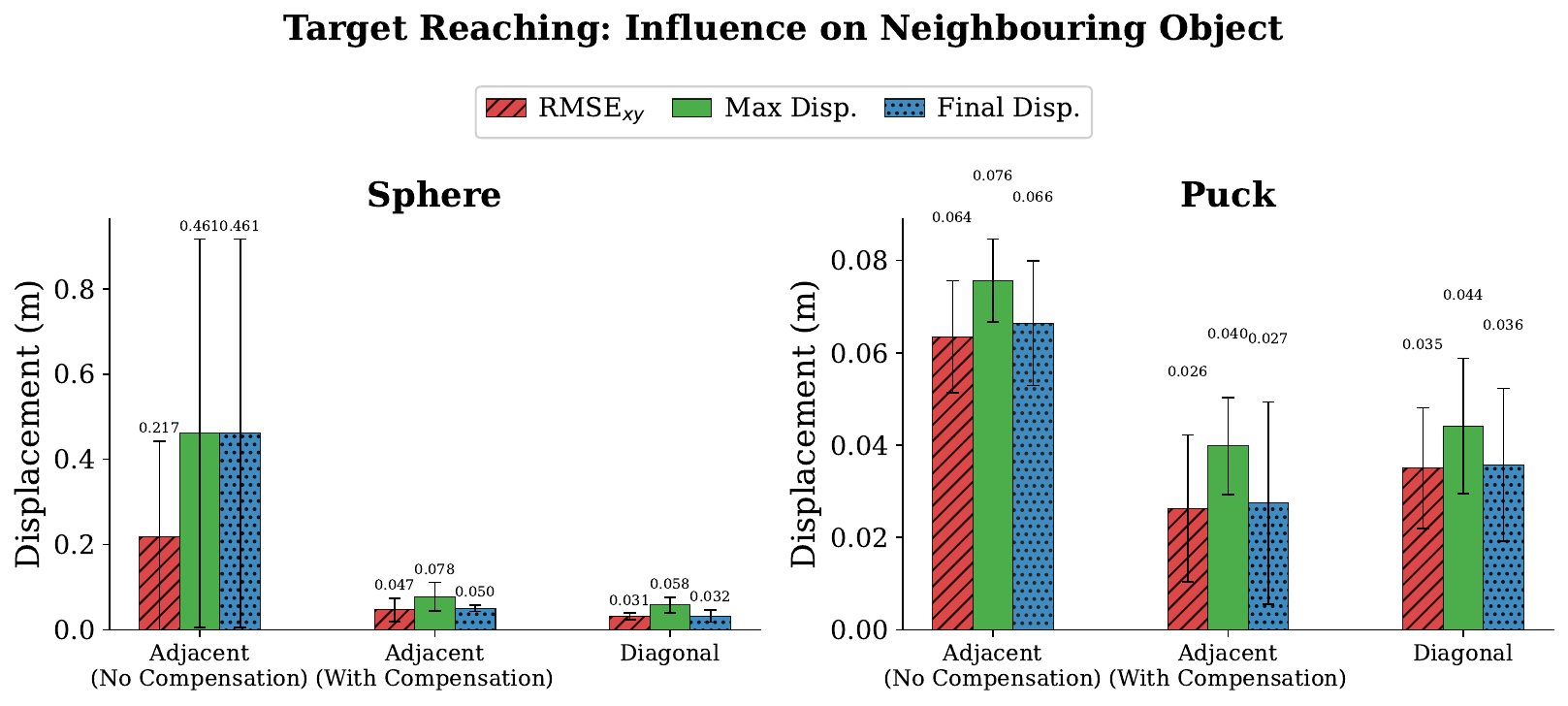}
    \caption{Target reaching: influence on a stationary
    neighbouring object. Three displacement metrics reported for
    adjacent (No compensation and compensation) and diagonal configurations.
    Left: sphere. Right: puck. Stabilised adjacent performance
    approaches diagonal levels across all metrics.}
    \label{fig:influence_target}
\end{figure}

\begin{figure}
    \centering
    \includegraphics[width=\linewidth]{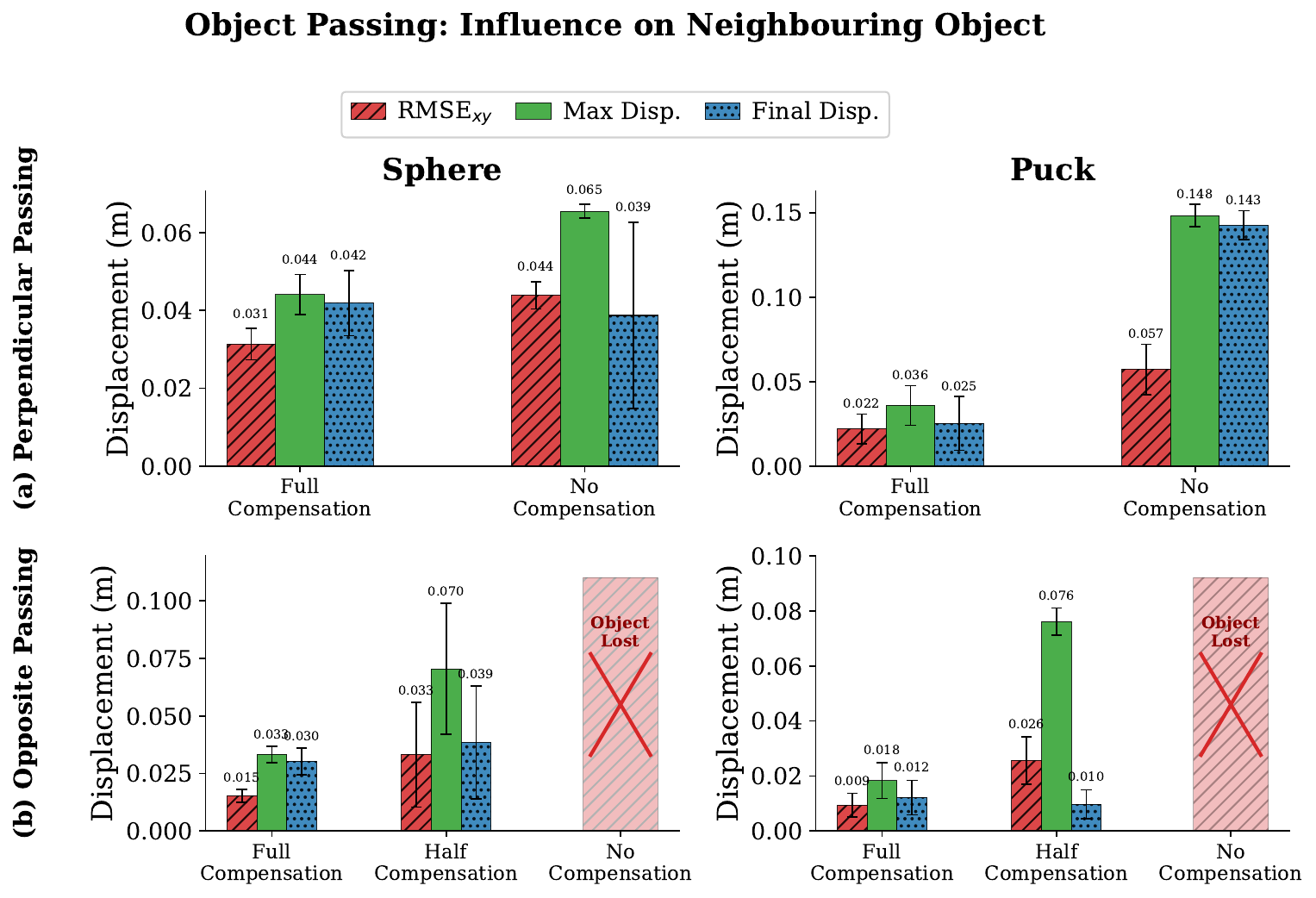}
    \caption{Object passing: influence on a stationary
    neighbouring object. (a)~Perpendicular passing (one shared
    actuator active). (b)~Opposite passing (both shared actuators
    active); red crosses indicate object loss without stabilisation.
    Left: sphere. Right: puck.}
    \label{fig:influence_passing}
\end{figure}

A key consequence of shared-boundary actuation is that
manipulating an object on one module perturbs the surfaces of
its neighbours. To quantify this interference, we measure the
displacement of a stationary object on a passive module while
another object is actively manipulated on an adjacent module
(Fig.~\ref{fig:interference_schematic}). We report three
complementary metrics: $\text{RMSE}_{xy}$ (average displacement
over the manipulation duration), maximum displacement
$d_{\max}$ (worst-case perturbation), and final displacement
$d_{\text{final}}$ (permanent drift after manipulation ends). A
sphere and puck serve as stationary objects, representing
rolling-dominant and sliding-dominant geometries respectively.
Two sources of interference are evaluated: target reaching and
object passing. All the experiments are repeated three times. 

% A key consequence of shared-boundary actuation is that
% manipulating an object on one module perturbs the surfaces of
% its neighbours. To quantify this interference, we measure the
% displacement of a stationary object on a passive module while
% another object is actively manipulated on an adjacent module. We
% report three complementary metrics: $\text{RMSE}_{xy}$ (average
% displacement over the manipulation duration), maximum
% displacement $d_{\max}$ (worst-case perturbation), and final
% displacement $d_{\text{final}}$ (permanent drift after
% manipulation ends). A sphere and puck serve as stationary
% objects, representing rolling-dominant and sliding-dominant
% geometries respectively, since interference sensitivity depends
% strongly on contact characteristics. Two sources of interference
% are evaluated: target reaching and object passing.

\subsubsection{Target Reaching Influence}

We examine two spatial configurations
(Fig.~\ref{fig:interference_schematic}a,b): \textit{adjacent
modules}(4-neighbour), where two actuators are shared between active and
passive modules (e.g., A3 and A4 between M0 and M2), and
\textit{diagonal modules}(diagonal-neighbour), where only a single actuator is
shared (A4 between M0 and M3).

% We examine two spatial configurations on the 2$\times$2
% platform: \textit{adjacent modules} (4-neighbour), where two
% actuators are shared between the active and passive modules
% (e.g., A3 and A4 between M0 and M1), and \textit{diagonal
% modules} (diagonal-neighbour), where only a single actuator is
% shared (A4 between M0 and M3). For each configuration, target
% reaching is performed on the active module while tracking the
% stationary object's displacement on the passive module.

Without compensation, the adjacent configuration produces
severe sphere interference ($\text{RMSE}_{xy}$: 0.217~m,
$d_{\max}$: 0.460~m, $d_{\text{final}}$: 0.461~m), with the
near-equal $d_{\max}$ and $d_{\text{final}}$ confirming
permanent drift and frequent module exit
(Fig.~\ref{fig:influence_target}). The puck shows lower but
substantial sensitivity ($\text{RMSE}_{xy}$: 0.064~m,
$d_{\max}$: 0.076~m) due to its flat contact geometry resisting
rolling. The diagonal configuration exhibits markedly less
coupling (sphere $\text{RMSE}_{xy}$: 0.031~m, puck: 0.035~m),
confirming that interference scales with the number of shared
actuators.

% Without stabilisation (Unstable), the adjacent configuration
% produces severe sphere interference ($\text{RMSE}_{xy}$:
% 0.217~m, $d_{\max}$: 0.460~m, $d_{\text{final}}$: 0.461~m),
% with the near-equal $d_{\max}$ and $d_{\text{final}}$ confirming
% that the sphere drifts permanently and frequently exits the
% module (Fig.~\ref{fig:influence_target}). The puck shows lower
% but still substantial sensitivity ($\text{RMSE}_{xy}$: 0.064~m,
% $d_{\max}$: 0.076~m) due to its flat contact geometry resisting
% rolling. The diagonal configuration exhibits markedly less
% coupling for both objects (sphere $\text{RMSE}_{xy}$: 0.031~m,
% puck: 0.035~m), confirming that interference scales with the
% number of shared actuators.

% To mitigate adjacent interference, we propose a
% \textit{stabilisation strategy}: the two non-shared actuators (Red) of
% the passive module are raised to 0.35~m, creating a counter-tilt
% that maintains a locally level surface when shared actuators (A3 and A4) are
% moved by the M0 module. This reduces sphere
% $\text{RMSE}_{xy}$ to 0.047~m (78\% reduction) and puck to
% 0.026~m (59\% reduction), approaching diagonal-level
% performance (Fig.~\ref{fig:influence_target}).

To mitigate adjacent interference, we propose a
\textit{stabilisation strategy}: the two non-shared actuators (Fig \ref{fig:interference_schematic}, red actuators) of
the passive module are raised to 0.35~m. When
shared actuators (A3 and A4) are moved during active manipulation on the
M0 module, the passive module's M2 surface develops an
unintended tilt toward the active module. Raising the non-shared
actuators creates a counter-tilt that maintains a locally level
surface, preventing object displacement. This reduces sphere
$\text{RMSE}_{xy}$ to 0.047~m (78\% reduction), $d_{\max}$ to
0.078~m, and $d_{\text{final}}$ to 0.050~m, approaching
diagonal level performance. The puck similarly improves to
$\text{RMSE}_{xy}$: 0.026~m (59\% reduction), $d_{\max}$:
0.040~m (Fig.~\ref{fig:influence_target}).

\subsubsection{Object Passing Influence}

During inter-module transfer, shared actuators are actively
raised to create the passing slope, producing a distinct
interference profile compared to target reaching. Two passing directions are evaluated
relative to the passive module
(Fig.~\ref{fig:interference_schematic}c,d):

% During inter-module transfer, shared actuators are actively
% raised to create the slope that drives object motion, producing
% a distinct interference profile compared to target reaching. Two
% passing directions are evaluated relative to the passive module
% (Fig.~\ref{fig:influence_passing}):

\textbf{Perpendicular passing}
(Fig.~\ref{fig:interference_schematic}c): Object~1 is passed
between modules sharing one actuator with the passive module
(e.g., M0 to M1 while Object~2 rests on M2). Without
stabilisation, moderate interference occurs (sphere
$\text{RMSE}_{xy}$: 0.044~m; puck: 0.057~m). The puck exhibits
high $d_{\max}$ (0.148~m) and $d_{\text{final}}$ (0.143~m)
despite lower $\text{RMSE}_{xy}$, indicating that once
displaced, flat objects do not self-correct. Stabilisation where the non-shared
actuators of the passive module mirror the passing action
reduces sphere $\text{RMSE}_{xy}$ to 0.031~m and puck to
0.022~m (Fig.~\ref{fig:influence_passing}a).

% \textbf{Perpendicular passing:} Object~1 is passed between two
% modules that share one actuator with the passive module (e.g.,
% M0 to M1 while Object~2 rests on M2; only A3 is shared between
% the active path and the passive module). Without stabilisation,
% moderate interference occurs (sphere $\text{RMSE}_{xy}$:
% 0.044~m, $d_{\max}$: 0.065~m; puck $\text{RMSE}_{xy}$:
% 0.057~m). Notably, the puck exhibits high $d_{\max}$ (0.148~m)
% and $d_{\text{final}}$ (0.143~m) despite a lower
% $\text{RMSE}_{xy}$, indicating that once displaced, flat objects
% do not self-correct. Stabilisation where the non-shared
% actuators of the passive module mirror the passing action
% reduces sphere $\text{RMSE}_{xy}$ to 0.031~m and puck to
% 0.022~m (Fig.~\ref{fig:influence_passing}a).

% \textbf{Opposite passing}
% (Fig.~\ref{fig:interference_schematic}d): Object~1 is passed
% outward from M0, causing both shared actuators (A3, A4) to rise
% simultaneously. Without stabilisation, stationary object is lost from
% the passive module entirely. Full-mirror stabilisation achieves
% the lowest displacement (sphere: 0.015~m, puck: 0.009~m);
% half-mirror stabilisation yields intermediate values (sphere:
% 0.033~m, puck: 0.026~m) while preserving partial actuation
% freedom (Fig.~\ref{fig:influence_passing}b).

\textbf{Opposite passing:} (Fig.~\ref{fig:interference_schematic}d): Object~1 is passed
outward from M0, causing both shared actuators (A3, A4) to rise
simultaneously. Without compensation,
this produces critical interference, with displacement exceeding
the module boundary and resulting in complete object loss for
stationary object. Two mirroring strategies address this:
\textit{full-mirror compensation}, where non-shared actuators (red)
match the passing height (green actuators), achieves the lowest displacement
(sphere: 0.015~m, puck: 0.009~m); \textit{half-mirror
compensation}, at half the passing height, yields intermediate
values (sphere: 0.033~m, puck: 0.026~m) while preserving
partial actuation freedom on the passive module
(Fig.~\ref{fig:influence_passing}b).

Across all conditions (Table~\ref{tab:interference_summary}),
interference severity scales directly with the number of active
shared actuators, and spherical objects are consistently more
sensitive than flat objects during target reaching, while flat
objects show greater permanent drift during passing. The
proposed stabilisation strategies reduce $\text{RMSE}_{xy}$ by
30--78\% across configurations, confirming that the coupling
inherent to $(n+1)^2$ shared-boundary actuation can be
effectively managed through actuator compensation.

\begin{table}[t]
\centering
\caption{Interference summary across all conditions. All
displacement values in meters.}
\label{tab:interference_summary}
\resizebox{\columnwidth}{!}{%
\begin{tabular}{llccccccc}
\toprule
 & & \multicolumn{3}{c}{\textbf{No Compensation}} &
\multicolumn{3}{c}{\textbf{Full Compensation}} & \\
\cmidrule(lr){3-5} \cmidrule(lr){6-8}
\textbf{Scenario} & \textbf{Obj.} &
$\text{RMSE}$ & $d_{\max}$ & $d_{\text{fin}}$ &
$\text{RMSE}$ & $d_{\max}$ & $d_{\text{fin}}$ &
\textbf{$\Delta$RMSE} \\
\midrule
Target (Adj.) & Sph. & 0.217 & 0.460 & 0.461 & 0.047 & 0.078 & 0.050 & 78\% \\
Target (Adj.) & Puck & 0.064 & 0.076 & 0.066 & 0.026 & 0.040 & 0.027 & 59\% \\
Target (Diag.) & Sph. & 0.031 & 0.058 & 0.032 & -- & -- & -- & base \\
Target (Diag.) & Puck & 0.035 & 0.044 & 0.036 & -- & -- & -- & base \\
\midrule
Pass. (Perp.) & Sph. & 0.044 & 0.065 & 0.039 & 0.031 & 0.044 & 0.042 & 30\% \\
Pass. (Perp.) & Puck & 0.057 & 0.148 & 0.143 & 0.022 & 0.036 & 0.025 & 61\% \\
Pass. (Opp.) & Sph. & Lost & -- & -- & 0.015 & 0.030 & 0.030 & -- \\
Pass. (Opp.) & Puck & Lost & -- & -- & 0.009 & 0.018 & 0.012 & -- \\
\bottomrule
\end{tabular}%
}
\end{table}

%% file: discussion.tex
\section{Discussion}

The multi-modular platform demonstrates, for the first time,
that objects can be reliably transferred and precisely
positioned across interconnected soft fabric modules. The
hierarchical control framework achieves this without
learning-based methods, using only geometric deformation
principles and closed-loop feedback, notable given the fabric's
nonlinear catenary profile and friction-dependent dynamics. The
conflict-free Manhattan path planner provides a first layer of
interference management by ensuring that simultaneously
manipulated objects never occupy same modules. When
interference from adjacent operations is unavoidable, the
proposed compensation strategies reduce passive-object
displacement by 59--78\%, providing a second layer of mitigation
at the actuation level. The enabling shared-boundary
architecture covers a 1$\times$1~m workspace with 9 actuators
instead of the 16 that independent modules would require.
Beyond this practical hardware reduction, the architecture
fundamentally shapes the control problem: the interference study
shows that coupling scales directly with the number of shared
actuators, with adjacent modules (two shared) experiencing up to
0.217~m passive displacement while diagonal modules (one shared)
show only 0.031--0.035~m. This quantitative relationship
provides a predictive basis for larger configurations. In a
3$\times$3 grid, the central module shares actuators with four
neighbours, making interference management increasingly
critical. At larger scales, simultaneous manipulation across
multiple neighbouring modules can trigger cascading
displacements through chains of shared actuators. In such cases,
half-mirror compensation can contain the disturbance within a
few modules of the active site, as actuator perturbations
attenuate across successive shared boundaries and become
negligible after two to three modules.

Object geometry plays a dominant role across all experiments,
particularly during inter-module transfer. Spherical objects
show higher instantaneous sensitivity during target reaching
($\text{RMSE}_{xy}$: 0.217~m vs.\ 0.064~m for the puck), as
even small tilts induce rolling. However, during passing, flat
objects exhibit greater permanent drift ($d_{\text{final}}$:
0.143~m for puck vs.\ 0.039~m for sphere), indicating that once
displaced, flat objects lack a restoring mechanism. This
distinction suggests that object-aware control policies, even at
a coarse rolling-vs-sliding classification level, could enable
tailored compensation strategies for each module and inform the
choice of passing parameters during inter-module transfer.

The current system follows Manhattan-path trajectories aligned
to module grid axes. Finer within-module trajectory control is
constrained by the four-actuator configuration, which provides
only two independent tilt axes. Achieving arbitrary paths would
require either higher actuator density, contradicting the
reduced-density philosophy, or time-varying sequences exploiting
fabric dynamics, a promising future direction.

Regarding orientation control, the platform controls object
position but not orientation, relevant for high aspect-ratio
objects like cylinders and eggs that exhibited corrective zig-zag
motions during target reaching. Unlike rigid tilting platforms, the fabric
introduces position-dependent coupling between surface shape and
object pose. Potential approaches include exploiting asymmetric
friction during sequential tilt-and-release motions or rapid
oscillatory actuation for dynamic reorientation. 

While simulation closely mirrors the hardware and shows good
agreement at the 2$\times$2 scale, constructing a 3$\times$3
prototype remains a priority. The current inter-module transfer
strategy relies on open-loop timed actuation phases; closed-loop
transfer using real-time object tracking during the passing
phase itself could improve reliability for lighter or more
irregular objects. The interference study characterises coupling
for two objects; in larger configurations, multi-source
interference patterns become combinatorial, requiring
decentralised control strategies. Additionally, integrating
embedded sensing would reduce reliance on the external OptiTrack
system, and adaptive gain tuning based on object type and
interference conditions could further improve performance.

%% file: conclusion.tex
\section{Conclusion}

This work demonstrated, reliable
inter-module object transfer and coordinated manipulation across
interconnected soft fabric modules. A hierarchical control
framework combining conflict-free Manhattan-based path planning
with directional passing and geometric PID control achieves
sub-centimeter positioning accuracy, consistent transfer of
heterogeneous objects including fragile items, and parallel
multi-object manipulation. Interference from shared-boundary
actuation is managed through a two-layer approach: conflict-free
path planning prevents simultaneously active neighbouring
modules, while actuator compensation reduces residual
passive-object displacement by 59--78\%. The enabling
shared-boundary hardware architecture requires only $(n+1)^2$
actuators for an $n \times n$ grid, covering 1$\times$1~m with
9 actuators instead of 16 for independent modules. Together,
these findings confirm that the coupling inherent to
shared-boundary architectures can be effectively addressed
without sacrificing manipulation performance, establishing a
foundation for deploying soft manipulation surfaces in
large-area applications. Future work targets 3$\times$3 hardware
validation, closed-loop inter-module transfer, and embedded
sensing for object-aware manipulation. The structural parallels
with physically-coupled multi-agent systems suggest that the
platform can also serve as a testbed for studying decentralised
coordination in distributed manipulation.

%% file: ref.bib
@INPROCEEDINGS{11020841,
        author={Ingle, Pratik and Støy, Kasper and Faiña, Andres},
        booktitle={2025 IEEE 8th International Conference on Soft Robotics (RoboSoft)}, 
        title={Soft Manipulation Surface With Reduced Actuator Density For Heterogeneous Object Manipulation}, 
        year={2025},
        volume={},
        number={},
        pages={1-8},
        doi={10.1109/RoboSoft63089.2025.11020841}}

@INPROCEEDINGS{11163787,
  author={Ingle, Pratik and Støy, Kasper and Faiña, Andres},
  booktitle={2025 IEEE 21st International Conference on Automation Science and Engineering (CASE)}, 
  title={Heterogeneous object manipulation on nonlinear soft surface through linear controller}, 
  year={2025},
  volume={},
  number={},
  pages={1780-1787},
  doi={10.1109/CASE58245.2025.11163787}}

@inproceedings{todorov2012mujoco,
  title={MuJoCo: A physics engine for model-based control},
  author={Todorov, Emanuel and Erez, Tom and Tassa, Yuval},
  booktitle={2012 IEEE/RSJ International Conference on Intelligent Robots and Systems},
  pages={5026--5033},
  year={2012},
  organization={IEEE},
  doi={10.1109/IROS.2012.6386109}
}

@article{zhou2016controlling,
  title={Controlling the motion of multiple objects on a Chladni plate},
  author={Zhou, Quan and Sariola, Veikko and Latifi, Kourosh and Liimatainen, Ville},
  journal={Nature communications},
  volume={7},
  number={1},
  pages={12764},
  year={2016},
  publisher={Nature Publishing Group UK London}
}

@article{moon2006distributed,
  title={Distributed manipulation of flat objects with two airflow sinks},
  author={Moon, Hyungpil and Luntz, Jonathan},
  journal={IEEE Transactions on robotics},
  volume={22},
  number={6},
  pages={1189--1201},
  year={2006},
  publisher={IEEE}
}

@article{uriarte2022methode,
  title={Methode zur Bewertung der Flexibilit{\"a}t und Wandelbarkeit am Beispiel eines omnidirektionalen F{\"o}rdersystems},
  author={Uriarte, Claudio and Thamer, Hendrik},
  journal={Logistics Journal: Proceedings},
  volume={2022},
  number={18},
  year={2022}
}

@inproceedings{xue2024arraybot,
  title={Arraybot: Reinforcement learning for generalizable distributed manipulation through touch},
  author={Xue, Zhengrong and Zhang, Han and Cheng, Jingwen and He, Zhengmao and Ju, Yuanchen and Lin, Changyi and Zhang, Gu and Xu, Huazhe},
  booktitle={2024 IEEE International Conference on Robotics and Automation (ICRA)},
  pages={16744--16751},
  year={2024},
  organization={IEEE}
}

@article{johnson2023multifunctional,
  title={A multifunctional soft robotic shape display with high-speed actuation, sensing, and control},
  author={Johnson, BK and Naris, M and Sundaram, V and Volchko, A and Ly, K and Mitchell, SK and Acome, E and Kellaris, N and Keplinger, C and Correll, N and others},
  journal={Nature Communications},
  volume={14},
  number={1},
  pages={4516},
  year={2023},
  publisher={Nature Publishing Group UK London}
}

@inproceedings{yu2008morpho,
  title={Morpho: A self-deformable modular robot inspired by cellular structure},
  author={Yu, Chih-Han and Haller, Kristina and Ingber, Donald and Nagpal, Radhika},
  booktitle={2008 IEEE/RSJ International Conference on Intelligent Robots and Systems},
  pages={3571--3578},
  year={2008},
  organization={IEEE}
}
